\documentclass[10pt,twocolumn,letterpaper]{article}
\usepackage[pagenumbers]{cvpr} % To force page numbers, e.g. for an arXiv version

\usepackage{graphicx}
\usepackage{amsmath}
\usepackage{amssymb}
\usepackage{booktabs}
\usepackage{multirow}
\usepackage[usestackEOL]{stackengine}
\usepackage[toc,page,header]{appendix}
\usepackage{minitoc}
\usepackage[accsupp]{axessibility}  % Improves PDF readability for those with disabilities.
\usepackage{url}
\usepackage{booktabs}       % professional-quality tables
\usepackage{amsfonts}       % blackboard math symbols
\usepackage{nicefrac}       % compact symbols for 1/2, etc.
\usepackage{microtype}      % microtypography
\usepackage{xcolor}         % colors
\usepackage{graphicx}
\usepackage{caption}
\usepackage{subcaption}
\usepackage{amsmath}
\usepackage{amsfonts}
\usepackage{booktabs}
\usepackage{siunitx}
\usepackage{wrapfig}
\usepackage[pagebackref,breaklinks,colorlinks]{hyperref}

\newcommand{\upd}[1]{{#1}}
\usepackage[capitalize]{cleveref}
\crefname{section}{Sec.}{Secs.}
\Crefname{section}{Section}{Sections}
\Crefname{table}{Table}{Tables}
\crefname{table}{Tab.}{Tabs.}

\def\cvprPaperID{11191} % *** Enter the CVPR Paper ID here
\def\confName{CVPR}
\def\confYear{2022}

\begin{document}

\title{Everything at Once -- Multi-modal Fusion Transformer for Video Retrieval}
%\title{Everything at Once - Multimodal Fusion Transformer for Video Retrieval}

% \and

\author{%
    Nina Shvetsova $^1$ \quad
    Brian Chen $^2$  \quad
    Andrew Rouditchenko$^3$  \quad
    Samuel Thomas$^{4,5}$  \\
    % \vspace{1mm} \\
    Brian Kingsbury$^{4,5}$ \quad 
    Rogerio Feris$^{4,5}$ \quad 
    David Harwath$^6$  \quad 
    James Glass$^3$ \quad 
    Hilde Kuehne$^{1,5}$ \\
    % \vspace{1mm} \\
    \small{
    $^1$Goethe University Frankfurt,   
    $^2$Columbia University, 
    $^3$MIT CSAIL
    $^4$IBM Research AI, 
    $^5$MIT-IBM Watson AI Lab,
    $^6$ UT Austin
    % \vspace{2mm}
    } \\
    \small{
    \texttt{shvetsov@uni-frankfurt.de}}  
}

\maketitle

%%%%%%%%% ABSTRACT

\begin{abstract}
Multi-modal learning from video data has seen increased attention recently as it allows training of semantically meaningful embeddings without human annotation, enabling tasks like zero-shot retrieval and action localization. 
In this work, we present a multi-modal, modality agnostic fusion transformer that learns to exchange information between multiple modalities, such as video, audio, and text, and integrate them into a fused representation in a joined multi-modal embedding space.
We propose to train the system with a combinatorial loss on everything at once \upd{-- any combination of input modalities, such as} single modalities as well as pairs of modalities, explicitly leaving out any add-ons such as position or modality encoding.
At test time, the resulting model can process and fuse any number of input modalities. Moreover, the implicit properties of the transformer allow to process inputs of different lengths. 
To evaluate the proposed approach, we train the model on the large scale HowTo100M dataset and evaluate the resulting embedding space on four challenging benchmark datasets obtaining state-of-the-art results in zero-shot video retrieval and zero-shot video action localization. 
\upd{Our code for this work is also available.\footnote{\url{{https://github.com/ninatu/everything_at_once}}}}

\end{abstract}

%%%%%%%%% BODY TEXT]

\section{Introduction}
\label{sec:intro}

\begin{figure}[t]
    \centering 
    \vspace{0.3cm}
    \begin{subfigure}{\linewidth}
        \includegraphics[width=\textwidth]{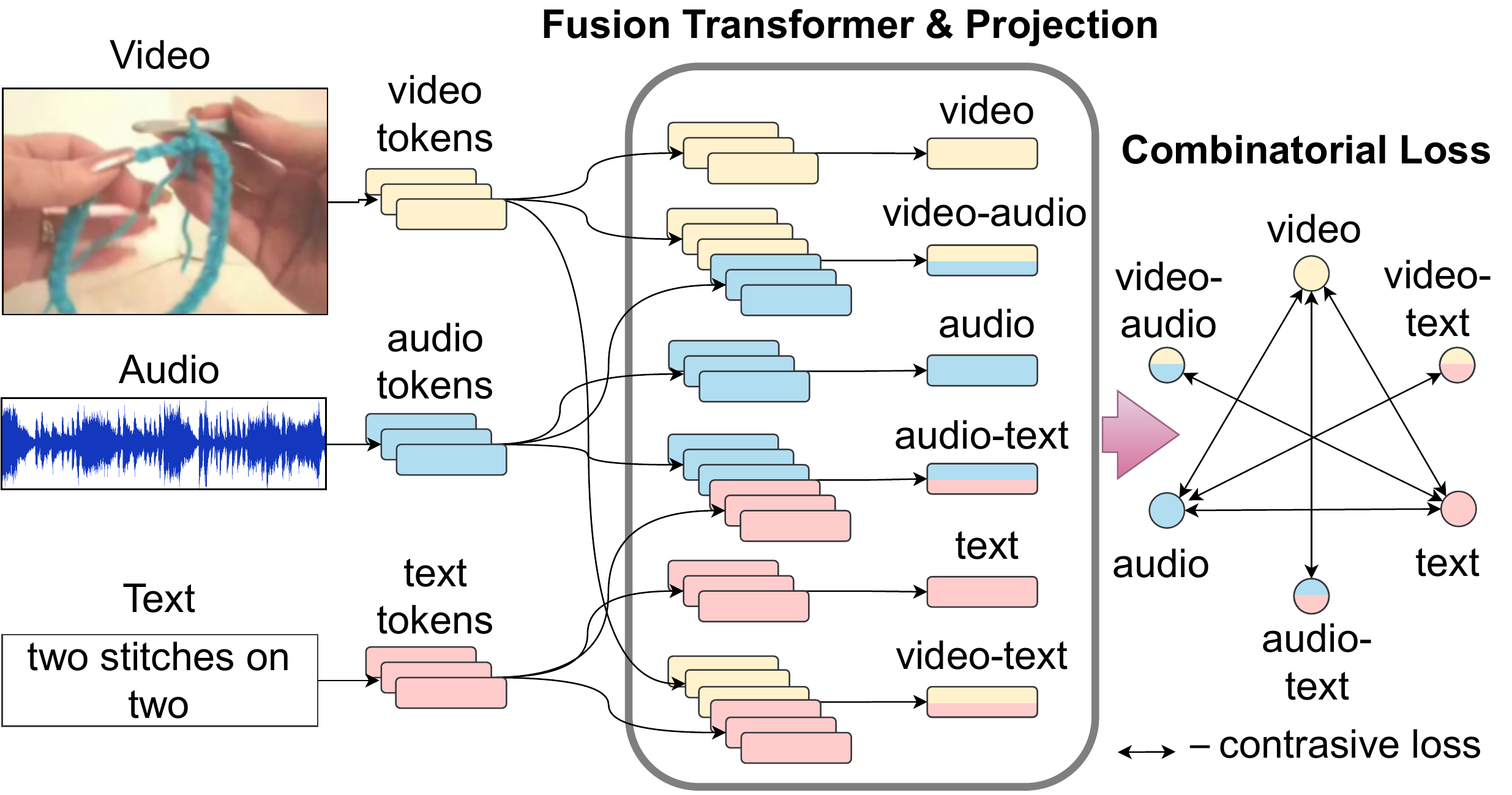}
        % \captionof{figure}{}
    \end{subfigure}
    % \vfill
    \vspace{-0.1cm}
    \caption{Overview of the proposed approach for self-supervised learning of multi-modal embedding space. The fusion transformer is able to process any combination of input modalities. Internally, the transformer allows each modality to attend to each other. The proposed architecture is trained with a combinatorial contrastive loss considering each possible combination of input modalities.
    \label{fig:diagram}}
    \vspace{-0.5cm}
\end{figure}

Humans capture their world in various ways, combining different sensory input modalities such as vision, sound, touch, and more, to make sense of their environment.
Video data approximates this type of input by combining visual and audio information as two coherent and complementary signals that can be further enhanced with a text description. 
Recent research has therefore started to explore how the information of those different modalities can be leveraged to learn meaningful representations from this kind of content. 
Such systems can be used for representation learning, for example, to learn multi-modal embedding spaces on video data 
\cite{alayrac2020self, Akbari2021VATT},
where the input of one modality such as text, can be matched to one or more other modalities such as video and audio, enabling tasks such as nearest-neighbor based zero-shot classification or video retrieval
\cite{miech2019howto100m,gabeur2020multi,rouditchenko2020avlnet}.
Our work in this paper focuses on the later problem, namely the learning of meaningful multi-modal embedding spaces. 
Current approaches in this area usually learn encodings for different modalities by projecting inputs to a common space \upd{and applying contrastive loss to bring embeddings from co-occurred modalities together.}
% and computing the similarity by using the dot product between the projected embeddings.
Such approaches can be based on classical neural network elements to learn those encodings~\cite{miech2019howto100m, rouditchenko2020avlnet, chen2021multimodal,miech2020end,amrani2020noise}, i.e. 
\upd{convolutional neural networks backbones and non-linear projections}~\cite{miech2019howto100m},
multiple instance learning \cite{miech2020end}, or
clustering \cite{chen2021multimodal}.
% using a combination of fully connected linear layers and multiple instance learning \cite{miech2020end}, gating \cite{rouditchenko2020avlnet}, clustering \cite{chen2021multimodal}. 
More recently transformer based methods have also been proposed~\cite{gabeur2020multi,Bain2021Frozen, Akbari2021VATT,Luo2021CLIP4Clip}. To generate the final embedding space, they use multiple independent single-modality self-attention transformer blocks~\cite{Bain2021Frozen,Luo2021CLIP4Clip,Guzhov2021AudioCLIP}, or a single transformer model for all modalities \cite{gabeur2020multi}, or a single modality-agnostic transformer~\cite{Akbari2021VATT}. In the last approach, modalities are still processed independently and one-by-one forwarded to achieve a single-modality embedding. 
But so far, none of these transformers allow for adaption to any given number of input modalities. 
Although modality-agnostic transformers that handle multiple input modalities such as PerceiverIO~\cite{Jaegle2021PerceiverIO} have been proposed, they have been constructed 
% with a different goal of 
for learning a latent space that can cover multiple tasks in different domains. Compared to our work, the latent space in such cases mainly serves the purpose of compressing multiple inputs and tasks in one model.

%While the architecture might enable information exchange between any combination of modalities, the focus is rather to enforce performance on different tasks, and thus might keep representations separate instead of joint.
%Finally, cross-modal transformer layers that process joint vision-text input have been well established in context of vision language architectures, as e.g. described in \cite{Khan2021TransformersInVision}, but so far have not been tried on more than two modalities.

% In this work, we propose an approach that leverages self-attention for multi-modal learning based on more than two inputs, allowing modalities can attend each other. 
In this work, we propose an approach that leverages self-attention for multi-modal learning which jointly processes any number of modalities and allows modalities to attend to each other. 
%To this end, our architecture combines elements of both classical and transformer architecture. 
A high level overview our architecture is shown in Figure~\ref{fig:diagram}.
\upd{Input tokens from one or more modalities 
are passed through a fusion transformer that attends features relevant for a combined input, followed by a projection to a joint multi-modal embedding space.}
% Input tokens from one or more modalities obtained by pre-trained backbones are passed trough a fusion transformer layer to attend features relevant for single modality as well as for combined input. The output tokens are then combined according to their modality and finally projected to a joint embedding space via a linear gating function. 
%The transformer only has to learn which features to attend, given a input of random modalities, to improve the joint embedding space learning in general. 
We design and train the fusion transformer to cover three aspects of multi-modal video learning: first, it should allow modalities to attend to each other and learn multi-modal correlations; second, it should be modality-agnostic and handle any possible modality input combination; and third, as different modalities and samples can vary in length, it should be able to process input of any length.
%
%Looking a cross modal processing in transformers, two main techniques have been established so far, universal attention, where all tokens are attend by all queries \cite{chen2020uniter} and co-attentional architecture as proposed in \cite{Lu2019ViLBERT}, where queries from one modality only attend to the other modality.
To enable the fusion transformer to address all those tasks, we follow the idea of a universal self-attention in the transformer block and share key, query, and value weights to all tokens, agnostic of their input modality. 
In this way, self-attentions learns which input tokens to attend from single modalities as well as from any combination of modalities in a general way.
% as shown in more detail in Figure~\ref{fig:diagram}.
%

To train the model, we propose a combinatorial loss function which considers contrastive loss between all possible and available input combinations.
For example, in the case of vision, text, and audio, the loss is based on each modality embedding alone as well as based on pairwise vision-text, audio-text, and text-audio combinations as shown in Figure~\ref{fig:diagram}.
%To allow for a balanced training we also propose to have a weight matrix for each possible loss combination to adapt the training to the characteristics of the targeted training or testing scenarios.
%
The resulting model is thus able to fuse any number of input modalities at test time.
% Compared to other universal self-attention methods, we omit any meta information encoding such as position or modality embedding. This further allows us to leverage the natural property of the transformer to process any input of different lengths, i.a. as we are no longer bound to a maximum input size defined at training time.
\upd{Compared to other universal self-attention methods, we omit any meta information encoding such as position or modality embedding. This further allows us to process any input of different lengths, as we are no longer bound to a maximum input size defined at training time.}
Note that while we refer to this transformer as a fusion transformer, we are not proposing a new transformer architecture, but rather refer to it as a transformer that is trained in a way that enables fusion without any need for changes to the self-attention mechanism.
% Practically, at training time, we input fix length segments, but at test time, we input any reasonable length.
As a result, the final modal can be used for any type of input, single modalities or combinations of multiple ones, as well as for any input length.
% and without any ordering constrains. 

%To this end, we also extend to the idea of a classical dual modality cross-modal transformer layer, as used in vision-language tasks, to more than two input modalities. 

%Compared to the alternative of having a co-attentional architecture, using a universal attention has, especially in context of more than two modalities, the advantage that the network does not need to grow exponentially with the number of modalities, as this might be the case for co-attention or size or the need for different networks. \hkc{doubling, fix later}   

We evaluate the proposed approach by training the model on the HowTo100M dataset \cite{miech2019howto100m} and testing its zero-shot text-to-video retrieval and step action localization on four downstream datasets, namely YouCook2~\cite{zhou2018towards}, MSR-VTT~\cite{xu2016msr}, CrossTask~\cite{zhukov2019cross} and Mining YouTube~\cite{kuehne2019mining}. 
% To be comparable with previous work and for resource efficiency, we use pre-extracted features from \cite{miech2019howto100m} and use those as input for our model. 
% Our results show that the simple combination of a transformer layer together with a linear embedding space learning is already able to improve current state-of-the-art methods across all tasks, even compared to other transformer-only based methods. Beyond that, 
\upd{Our results show that the proposed combination of a fusion transformer together with a combinatorial loss function improves performance and leads to new state-of-the-art results.} %Beyond that it shows that the proposed the combination of a fusion transformer layer together with a combinatorial loss function and the ability to process input any size again improves perfromace and leads to new state-of-the art results in the field.
We summarize the contributions of the paper as follows:
\begin{itemize}
    \item We propose a multi-modal fusion transformer that processes input of any combination of modalities and any length and attends relevant features with respect to cross-modal information.
    \item We propose a combinatorial contrastive loss that considers all possible combinations of input modalities at training time.
    \item We show that using such a multi-modal fusion transformer as an intermediate processing step can significantly improve performance for multi-modal embedding space learning.
\end{itemize}

%we present a framework that learns how to exchange information between multiple modalities, such as video, audio, and text, and integrate it into the joined multimodal-modal representation. We utilize a single transformer that may take tokens of different modalities, allowing them to attend to each other and exchange information, to obtain an embedding that aggregates temporal and multi-modal information. After pretraining such a transformer only on singles or pairs of modalities,  The proposed modal, trained in a self-supervised way on the large scale HowTo100M dataset, allowed us to obtain state-of-the-art in zero-shot video retrieval and zero-shot video segmental on four downstream datasets. 

\section{Related Work}

\textbf{Multi-modal learning.}
The idea of learning from more than one modality can be seen as an integral part of machine learning research, comprising areas such as vision-language learning \cite{Radford2021CLIP, Zhang_2021_VinVL}, vision-audio learning~\cite{Chen2020ICASSP, Xiao2020Audiovisual,Tsiami_2020_STAViS,arandjelovic2018objects,arandjelovic2017look,aytar2016soundnet,harwath2018jointly}, zero-shot learning~\cite{Huynh_2020_Shared, Mancini_2021_OpenWorld},  \upd{cross-modal generation~\cite{zhou2018visual,ma2019unpaired,reed2016generative}, as well as multi-modal multi-task learning~\cite{kaiser2017one}.} Video naturally combines multiple modalities, while at the same time allowing to learn from large-scale data that would not be annotatable in a reasonable time. In this context, Miech~et~al. \cite{miech2019howto100m} proposed the HowTo100M dataset of narrated videos and presented a system showing the potential of multi-modal learning for learning a video-text embedding space via contrastive loss. The dataset contains YouTube instructional videos that come with audio and respective subtitles as a textual description obtained by Automatic Speech Recognition (ASR). As this data can be considered more noisy than curated vision-text datasets, Amrani~et~al. \cite{amrani2020noise} proposed a noise estimation for multi-modal data via multi-modal density estimation. 
% As an extension of that, 
Miech~et~al. \cite{miech2020end} proposed MIL-NCE, combining the idea of noise-contrastive estimation with a multiple instance learning formulation. Alwassel~et~al. \cite{alwassel_2020_xdc} used the audio and video information only and proposed 
% a cross-modal deep clustering framework
to leverage unsupervised clustering as a supervisory signal across modalities.
While those works \cite{miech2019howto100m, amrani2020noise, miech2020end, alwassel_2020_xdc} only use two modalities to train their models, others have focused on the problem of learning from vision, audio, and text at once~\cite{Aytar2017See,alayrac2020self,duarte2021routing,rouditchenko2020avlnet,chen2021multimodal}. As perhaps one of the first, Aytar~et~al. \cite{Aytar2017See} proposed an architecture trained on image-text and image-audio pairs that allows to connect text and audio modalities. Later Alayrac~et~al. \cite{alayrac2020self} followed the idea of different embedding spaces for different modality combinations and proposed Multi-Modal Versatile Networks.
% with a separation of fine and coarse spaces. 
% with up to seven embedding networks to cross the bridge between all three modalities. 
A shared embedding space was proposed by Rouditchenko~et~al. \cite{rouditchenko2020avlnet} mapping all three modalities in one joint space.
% via a combination of gating and contastive max margin softmax loss. 
This idea has recently been extended by additional clustering and reconstruction loss by Chen~et~al. \cite{chen2021multimodal}. 

%text-video-audio parers: See, Hear, and Read: Deep Aligned Representations
%Yusuf Aytar, Carl Vondrick, Antonio Torralba 

\textbf{Multi-modal learning with transformers.} Architectures based on self-attention and transformers have been explored to learn from multi-modal video data. Cheng~et~al. \cite{Cheng2020look} proposed a co-attention module to learn correspondences between audio and video samples. 
Luo~et~al. \cite{luo2020univilm} pick up on that idea but proposed, similar to Uniter \cite{chen2020uniter} for vision-language tasks, a joint cross-modal encoding for video-text pairs. 
Compared to that, Bain~et~al. \cite{Bain2021Frozen} focused on the problem of how to attend to temporal as well as spatial information in the video backbone. They therefore processed both modalities, video and text, in two separate transformer backbones and only added a linear mapping layer on top of the backbones. \upd{In this context, recently, Nagrani~et~al.~\cite{nagrani2021attention} proposed a multi-modal bottleneck transformer for effective audio-visual fusion trained in the supervised setting.}
A transformer-based approach that actually uses all three modalities, and can therefore be considered closest to our proposed work, has been proposed by Akbari~et~al. \cite{Akbari2021VATT}. Here, a single backbone transformer is applied to any of the modalities separately, but with shared attention. For training, the model follows the idea of \cite{alayrac2020self} and computes the matching of video-audio first, followed by video-text matching. It thus fuses those modalities in a pairwise way, which can be compared to a subset of our proposed loss function. 
Other approaches also leverage temporal aspects in context of multi-modal transformer learning. Gabeur~et~al. \cite{gabeur2020multi} used a combination of expert and temporal embeddings to train a multi-modal transformer while Wang~et~al. \cite{Wang_2021_T2VLAD} proposed a local-global temporal alignment based on multi-modal experts to guide the training. 
% To our best knowledge, so far no architecture has been proposed that uses all three modalities as a joint input for a self-attention layer and computes the respective contrastive loss over all possible fusion combinations. 
The idea of simply using a pretrained vision-language transformer model has also been explored by Lou~et~al. \cite{Luo2021CLIP4Clip}, using the pretrained CLIP model \cite{Radford2021CLIP} as a backbone with a transformer-based similarity encoder on top of a vision and text backbone and achieving good results on tasks such as video retrieval. As most transformer-based method use various and sometimes not-publicly available datasets for backbone pretraining or have a need for resources that make it hard to repeat experiments, it is difficult to directly compare performance across different architectures and pretraining set. 
We therefore decided to follow the setup used in majority of works here and rely on pre-extracted features that are then processed by the proposed architecture to allow a direct comparison with previous works.

% \paragraph{"Everything at Once" Model}

\section{Method}

\begin{figure*}[t]
    \centering 
    \begin{subfigure}{\linewidth}
        \includegraphics[width=\textwidth]{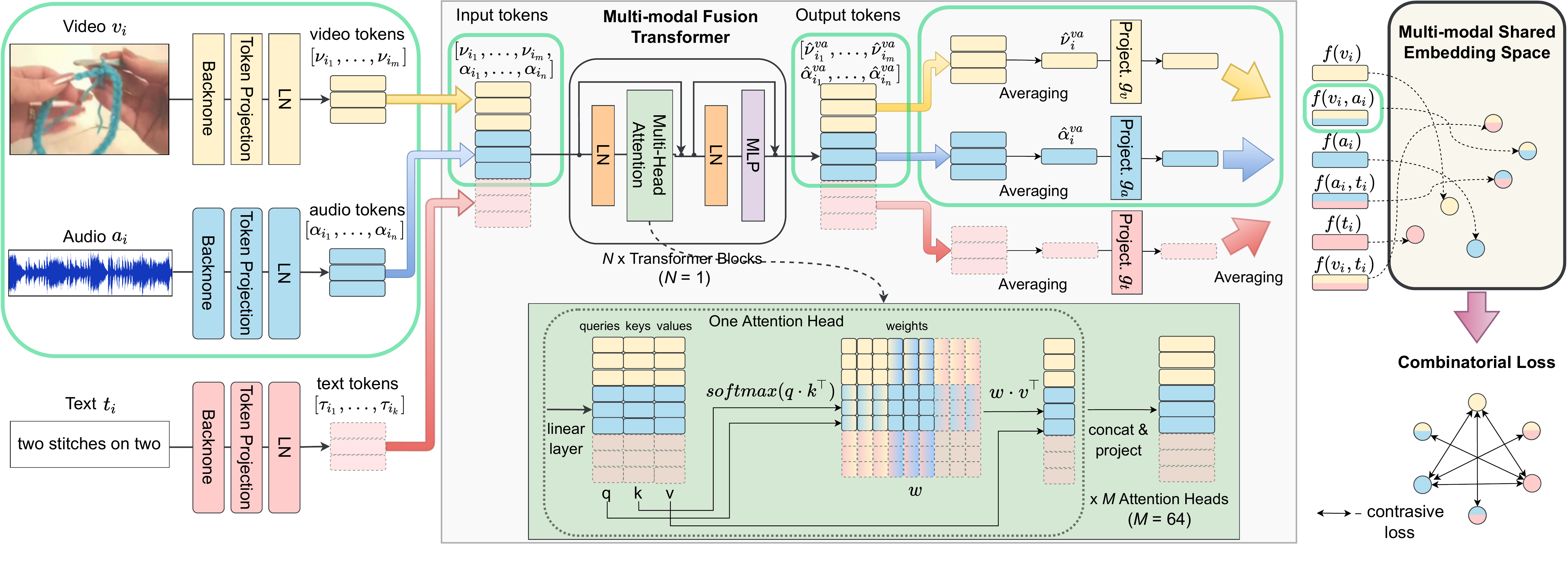}
        % \captionof{figure}{}
    \end{subfigure}
    % \vfill
    \vspace{-0.45cm}
    \caption{Schematic visualization the proposed method.  %showed all components of our model to extract a multi-modal embedding for any number of input modalities. Note that not all blocks are always applied, but only those that correspond to the given modalities: 
    While tokens from different modalities are processed in all possible combinations, we exemplary consider the video-audio pair marked with green rectangles here. The input tokens are forwarded together through the fusion transformer layer and attended by the respective weights, which are based on the combinations of keys and queries of input tokens in various modalities. The resulting outputs of multiple heads are then concatenated and projected to the final token space, which is then used to project each modality separately into the joint embedding space.  During training, we apply the model six times to obtain six embeddings corresponding to text, video, audio, text-video, text-audio, and video-audio modalities to compute the combinatorial loss.
    % Our model starts with token extraction from three modalities using pretrained backbone network, a projection layer, and normalization layer LN. 
    \label{fig:method}}
    \vspace{-0.25cm}
\end{figure*}

Our goal is to learn a projection function of single modalities or a set of modalities into a joint embedding space in a way that semantically similar inputs would be close to each other, e.g. the projection of the text description of a video scene should be close to the projection of the video-audio representation of this scene. In the following, we consider three modalities, video, audio, and text (corresponding ASR caption or linguistic narration); but the proposed method can be extended to more modalities. 
%   \nina{can be deleted:} Following previous works~\cite{rouditchenko2020avlnet,chen2021multimodal,gabeur2020multi,miech2020end,miech2019howto100m,alayrac2020self}, we train our multi-modal model in a self-supervised fashion with unlabeled instructional videos but using frozen pre-trained backbones for feature extraction.
%
%While instructional videos usually come with video and audio streams, the text is obtained by applying an automated speech recognition (ASR) system to the audio track. 
%Given a set of clips, the multi-modal shared space is trained by utilizing video-audio-text co-occurrence in the clips by applying contrastive loss to the dot-product similarities of obtained representations. Our goal is not only  to learn a mapping of a single modalities into the joint embedding space but also a mapping of any combination of input modalities, which is realized via the proposed fusion transformer and combinatorial loss. 

%The rest of the section is organized in the following way: Section~\ref{sec:problem} describes the problem statement more formally, Section~\ref{sec:model} introduces the architecture of our model, and Section~\ref{sec:loss} proposes our combinatorial loss.

\subsection{Problem Statement}
\label{sec:problem}

Given a set of text-video-audio triplets $\{(t_i, v_i, a_i)\}_{i=1}^N \in (\mathcal{T} \times \mathcal{V} \times \mathcal{A})^N$ of $N$ video clips obtained from the data distribution, 
% $P(\mathcal{T} \times \mathcal{V} \times \mathcal{A})$,
we are learning a projection $f(\cdot, \cdot, \cdot)$ that can take up to three inputs: text $t$, video $v$, and audio $a$ and produce $d-$dimensional embedding representation of the input. For the simplicity of the notation, we will omit missing modalities, so that  $f(t, v)$ will stand for projection $\mathcal{T} \times \mathcal{V} \to \mathbb{R}^d$ and represent the joint embedding of text $t$ and video $v$.
% , or $f(a)$ will stand for projection $\mathcal{A} \to \mathbb{R}^d$ and represent an embedding of the audio $a$. 
Our goal is to maximize the dot-product similarities between semantically related inputs  $f(t)$, $f(v)$, $f(a)$, $f(t, a)$, $f(t, v)$, $f(v, a)$ (such as when $t$, $v$, and $a$ are from the same video clip) and minimize otherwise.

\subsection{Model Architecture}

\label{sec:model}

\subsubsection{Token Creation}

As illustrated in Figure~\ref{fig:method}, our architecture starts from features extracted from modality-specific backbones. 
%To better aggregate and exchange single modality or cross-modal information in a multi-modal transformer network, we are leveraging local-temporal relations in data. To this end, we do not pool features as~\cite{miech2019howto100m,miech2020end,rouditchenko2020avlnet,chen2021multimodal}, but 
We transform sets of extracted feature vectors into token space by learnable modality-specific projections and modality-specific normalization layers~\cite{ba2016layer}. As a result, for the $(t_i, v_i, a_i)$ input triplet, we obtain three sets of tokens: $[\tau_{i_1}, ..., \tau_{i_k}]$ from text $t_i$, $[\nu_{i_1}, ..., \nu_{i_m}]$ from video $v_i$, $[\alpha_{i_1}, ..., \alpha_{i_n}]$ from audio $a_i$.
%\subsubsection{Attention Masks}
As the number of tokens may vary, e.g. depending on the length of the video clip, we normalize the length of inputs per batch to allow batch processing by padding and using attention masks~\cite{vaswani2017attention}. 
% We pad sequences of tokens of each modality up to the largest length of this modality in a batch and create attention masks that indicate positions of padded tokens so that the model does not attend to them.  
Practically, for comparability, we follow the protocol of \cite{miech2019howto100m, rouditchenko2020avlnet} and train the model on fixed-length video clips.
% , so the attentions masks are mainly necessary only for the text input. 
Technically, the model can handle clips of variable length, also with respect to different modalities, at training and at test time if needed. 
%However, for testing, it allows us to handle clips of variable length, which is very important. 

\subsubsection{No Positional Embeddings} 

Unlike other transformer-based methods \cite{chen2020uniter,Bain2021Frozen, Akbari2021VATT,lei2021less,tang2021decembert}, we omit adding any positional or type embedding information to the tokens. The reason for this is three-fold. With respect to type embedding, it can be assumed that tokens already encode this information based on the fact that they are generated by different backbones, and hence each come with their own ``fingerprints.''
% Thus, tokens obtained from different modalities are already very different. 
%Regarding the tokens from the same modalities: intuitively, video streams as rbg-sequence can be easily ordered by humans: based on the knowledge that two neighboring frames are very similar, we can easily create a sequence one by one. The only direction of sequence can be challenging, but we can guess it by the context.  The same intuition work for the audio tokens and short texts. 
%Secondly, positional information is more beneficial with the longer pretraining with auxiliary tasks like masking. 
Positional information has been shown to be beneficial in the context of consistently structured data such as sentences. But in the case of multi-modal video learning, clips are sampled randomly from a larger video sequence at training time, usually without considering shot boundaries or speech pauses. We therefore do not expect a consistent temporal pattern in the sense that a clip always starts at the beginning of an action.
% or of a sentence.
Thus leaving out positional embedding might prevent adding noise during training. 
At inference time, avoiding the positional embedding allows us to process sequences longer than used in the training. % -- any possible length that fit in memory. 
% Since all other model blocks treat all tokens equally, having a positional embedding would be a limitation with respect to input length. 
% In our experiments we show that processing the whole video clip at once is more beneficial than cutting it into clips (Section~\ref{sec:ablation}).

\subsubsection{Multi-modal Fusion Transformer}

As our goal is to learn the representations of any number and combination of input modalities, we want the projection $f$ to learn how to fuse information from multiple modalities to enhance the joint embedding representation. For this purpose, we propose a multi-modal, modality agnostic transformer, where the keys, queries, and values of the input tokens and all further transformations are computed independently from the modality. 
To create our multi-modal fusion transformer, we adopt regular transformer blocks~\cite{vaswani2017attention}. Each transformer block consist of a multiheaded self-attention and a multilayer perceptron (MLP) with two LayerNorm (LN) transforms before them along with two residual connections, as illustrated in Figure~\ref{fig:method}. 
Note that the difference compared to other methods is not in the architecture itself, but in the way it is trained and the fact that the resulting fusion can actually be learned by a vanilla transformer block,  if it is specifically trained for this task. Fusion transformer thus refers to the way a transformer block can be used rather than to a new architecture. 

We train the system with a combinatorial input. Namely, we apply it to joint sets of input tokens from all possible combinations of modalities: singles - $t$, $v$, $a$ and pairs - $(t, v)$, $(v, a)$, $(t, a)$, allowing tokens from one modality to attend to tokens of other modalities. In this way, we can obtain a \textit{fused} representation from multiple modalities: the combination $(t, v)$ will result in a fused representation of text and video modalities denoted as $tv$, resp. for $va$ - video and audio, and $ta$ - text and audio. Note that e.g. in case of four modalities, we would consider all combinations up to a triplet $(t, v, a)$ during training. As more modalities would be added, the number of combinations would grow to the point where it might be infeasible to consider all configurations. In this case random modality dropout could be used during training as done in AVSlowfast~\cite{Xiao2020Audiovisual} or Perceiver~\cite{Jaegle2021PerceiverIO}.

% Our multi-modal Fusion Transformer is agnostic to the modalities and applied to any number of input modalities. 
Since we want the fusion transformer to be modality agnostic, in each training iteration, we apply it six times to obtain six representations for each sample $i$: $t_i$, $v_i$, $a_i$, $t_iv_i$, $v_ia_i$, $t_ia_i$. To obtain each representation, we create a joint list of tokens, e.g. for $v_ia_i$:  $[\nu_{i_1}, ..., \nu_{i_m}, \alpha_{i_1}, ..., \alpha_{i_n}]$. We apply the transformer to this input and obtain output tokens as e.g. $[\hat{\nu}^{va}_{i_1}, ..., \hat{\nu}^{va}_{i_m} \hat{\alpha}^{va}_{i_1}, ..., \hat{\alpha}^{va}_{i_n}]$ for $v_ia_i$ (with superscript $va$ denoting that tokens were attended to both $v$ and $a$ modalities), where each token was attended with information from another tokens. Note that, unlike the ViT model~\cite{dosovitskiy2020image}, we do not prepend a learnable $[cls]$ token, which usually serves as a joint representation of all tokens. In our ablation studies we show that this is beneficial for the model (Sec.~\ref{sec:ablation}).

\subsubsection{Projection to Shared Embedding Space}

With the resulting output tokens, we create the final embeddings for each modality. For each training sample, we get six output sets of tokens and thus embeddings. As an example, we  consider the case of creating the representation for $v_ia_i$. We divide output tokens $[\hat{\nu}^{va}_{i_1}, ..., \hat{\nu}^{va}_{i_m} \hat{\alpha}^{va}_{i_1}, ..., \hat{\alpha}^{va}_{i_n}]$ into groups based on modality: $[\hat{\nu}^{va}_{i_1}, ..., \hat{\nu}^{va}_{i_m}]$ and $[\hat{\alpha}^{va}_{i_1}, ..., \hat{\alpha}^{va}_{i_n}]$ and then average them: $\hat{\nu}^{va}_i = \sum_{j=1}^{m}{\hat{\nu}^{va}_{i_j}}$, $\hat{\alpha}^{va}_i = \sum_{j=1}^{n}{\hat{\alpha}^{va}_{i_j}}$. As a result, we obtain a vector representation for each modality included in this computation. %enhanced with information from others modalities. 
But since modalities, even when enhanced with other modalities, are still very different, we project them into the shared embedding space by the learnable modality-specific projections  $g_t$, $g_v$, or $g_a$ for projections for $t$, $v$, $a$ respectively, normalize them, and then combine into \textit{a final embedding vector}:
\begin{equation}
    f(v_i, a_i) = \textrm{norm}(\textrm{norm}(g_v(\hat{\nu}^{va}_i)) + \textrm{norm}(g_a(\hat{\alpha}^{va}_i))).
\end{equation}
The normalization (``norm'') is used to align the magnitude of vectors. When computing dot product similarity, we take into account only the angle between vectors.

\subsection{Combinatorial Loss}

\label{sec:loss}

\upd{Contrastive loss can be used to learn representations such that semantically similar inputs are mapped close to each other.} Unlike other methods~\cite{chen2021multimodal,Akbari2021VATT,alayrac2020self,rouditchenko2020avlnet} that learn how to bring modalities together by training with three pairwise single-modality contrastive losses, $L_{t\_v}$ between $(t, v)$, $L_{t\_a}$ between $(t, a)$, and $L_{v\_a}$ between $(v, a)$, we force tokens to exchange information between modalities while enabling additional contrastive losses: $L_{t\_va}$ between $(t, va)$, $L_{v\_ta}$ between $(v, ta)$, and $L_{a\_tv}$ between $(a, tv)$, and introduce our \textit{combinatorial loss}:
\begin{equation}
\begin{split}
\label{eq:loss}
    L = & \lambda_{t\_v}L_{t\_v} + \lambda_{v\_a}L_{v\_a} + \lambda_{t\_a} L_{t\_a} + \\
        & + \lambda_{t\_va} L_{t\_va} + \lambda_{v\_ta} L_{v\_ta} + \lambda_{a\_tv} L_{a\_tv},
\end{split}
\end{equation}
where $\lambda_{m\_\hat{m}}$ denotes a weighting coefficient of $(m, \hat{m})$.

\upd{Our combinatorial loss considers all possible and available modality combinations and can be generalized to any set of modalities $\mathcal{M} = \{m_1, ...,m_N\}$ as follows:
\begin{equation}
% \begin{split}
\label{eq:genloss}
L = \sum\limits_{\substack{
            \mathcal{X},\mathcal{Y} \subset \mathcal{M}; 
            \mathcal{X} \cap \mathcal{Y} = \varnothing}}  \lambda_{\mathcal{X}\mathcal{Y}} L_{\mathcal{X}\mathcal{Y}}.
% \end{split}
\end{equation}
where $L_{\mathcal{X}\mathcal{Y}}$ is a contrastive loss between the fused representations of subsets $\mathcal{X}$ and $\mathcal{Y}$, $\lambda_{\mathcal{X}\mathcal{Y}}$ is a weighting coefficient.}

To compute the contrastive losses for all combinations, we use Noise Contrastive Estimation~\cite{oord2018representation} with temperature $\tau$ and batch size $B$:
\begin{equation}
\begin{split}
        L_{\mathcal{X}\_\mathcal{Y}} &=
        -\frac{1}{B}\sum_{i=1}^B{\log\left(\frac{\exp(x_i^{\top}y_i/\tau)}{\sum_{j=1}^B{\exp({x_i}^{\top}{y_j}/\tau)}}\right)} - \\
        &- \frac{1}{B}\sum_{i=1}^B{\log\left(\frac{\exp(x_i^{\top}y_i/\tau)}{\sum_{j=1}^B{\exp({x_j}^{\top}{y_i}/\tau)}}\right)}.
\end{split}
\end{equation}

By combining both aspects, the processing of all possible modality combinations and the training of the system with the proposed combinatorial loss, we obtain a multi-modal fusion transformer that learns how to attend tokens from one modality to the tokens from all other modalities.

\section{Experimental Evaluation}

\begin{table*}
    % \tablestyle{3pt}{1.05}
    \resizebox{1\linewidth}{!}{
    \begin{tabular}{@{}l|c|c|c|c|ccc|cccc|cccc@{}}
    	\toprule
        \multirow{2}{*}{Method} & Train. & Retrieval & Train. & Visual & \multicolumn{3}{c}{Trainable BB} & \multicolumn{4}{c}{YouCook2} & \multicolumn{4}{c}{MSR-VTT} \\ 
    % 	\cmidrule(lr){4-7} \cmidrule(lr){8-11}
    	   & Mod. & & Dataset  & BB & $t$ & $v$ & $a$ & R@1$\uparrow$  & R@5$\uparrow$ & R@10$\uparrow$ & MedR$\downarrow$ & R@1$\uparrow$ & R@5$\uparrow$  & R@10$\uparrow$ & MedR$\downarrow$\\ 
    	\midrule
    % 	MIL-NCE~\cite{miech2020end} & HT100M & TV & I3D & word2vec & +(?) & (?) & (?) & 11.4 & 30.6 & 42.0 & 16 & 9.4 & 22.0 & 30.0 & 35 \\
    	ActBERT \cite{zhu2020actbert}  & $tv$ & $t \to v$ & HT100M  & Res3D+Faster R-CNN & & & & 9.6 & 26.7 & 38.0 & 19 & 8.6 & 23.4 & 33.1 & 36 \\
    	Support Set \cite{patrick2020support} & $tv$ & $t \to v$ & HT100M & R152 + R(2+1)D-34 & $\checkmark$ & & & - & - & - & - & 8.7 & 23.0 & 31.1 & 31 \\
    	HT100M \cite{miech2019howto100m} & $tv$ & $t \to v$ & HT100M & R152 + RX101 & & & & 6.1 & 17.3 & 24.8 & 46 & 7.5 & 21.2 & 29.6 & 38 \\
    	NoiseEstim. \cite{amrani2020noise} & $tv$ & $t \to v$ & HT100M & R152 + RX101 & & & & - & - &- & -& 8.4 & 22.0 & 30.4 & 36 \\
    	\upd{\textbf{Ours}} & $tv$ & $t \to v$ & HT100M & R152 + RX101 & & & & 11.2 & 28.5 & 39.7 & 19 & 9.6 & \textbf{26.1} & \textbf{36.1} & \textbf{23}\\
    	\textbf{Ours} & $tva$ & $t \to v$ & HT100M & R152 + RX101 & & & $\checkmark$ & 10.7 & 27.9 & 38.9 & 19 & 10.3 & 24.6 & 35.3 & 25\\
    	\midrule
    	MMT~\cite{gabeur2020multi} & $tva$ & $t \to va$ & HT100M & 7 experts  & $\checkmark$ & & & - & - & - & - & - & 14.4 & - & 66 \\
    	
    	AVLNet \cite{rouditchenko2020avlnet} & $tva$ & $t \to v+a$ & HT100M & R152+RX101 & & & $\checkmark$ & 19.9 & 36.1 & 44.3 & 16 & 8.3 & 19.2 & 27.4 & 47 \\
    	MCN\cite{chen2021multimodal} & $tva$ & $t \to v+a$ & HT100M & R152+RX101 & & & $\checkmark$ & 18.1 & 35.5 &  45.2 & - & \textbf{10.5} & 25.2 & 33.8 & - \\
    	\textbf{Ours} & $tva$ & $t \to va$ & HT100M & R152+RX101 & & & $\checkmark$ & \textbf{20.0} & \textbf{40.7} & \textbf{51.3} & \textbf{10} & 8.9 & 23.8 & 31.8 & 30  \\
    	\midrule
    	\multicolumn{16}{c}{\upd{Models with a stronger visual backbone:}} \\
    	\midrule
    % 	ActBERT \cite{zhu2020actbert}  & $tv$ & $t \to v$ & HT100M  & Res3D+Faster R-CNN & & & & 9.6 & 26.7 & 38.0 & 19 & 8.6 & 23.4 & 33.1 & 36 \\
    % 	Support Set \cite{patrick2020support} & $tv$ & $t \to v$ & HT100M & R152 + R(2+1)D-34 & $\checkmark$ & & & - & - & - & - & 8.7 & 23.0 & 31.1 & 31 \\
    	MMV \cite{alayrac2020self} & $tva$ & $t \to v$ & HT100M+AudioSet & TSM-50x2 & & $\checkmark$ & $\checkmark$ &11.7 & 33.4 & 45.4 & 13 & 9.3 & 23.0 & 31.1 & 38 \\
    	VATT \cite{chen2021multimodal} & $tva$ & $t \to v$ & AudioSet & Transformer & $\checkmark$ & $\checkmark$ & $\checkmark$ & - & - &  45.5 & 13 & - & - & 29.7 & 49 \\
    	MIL-NCE~\cite{miech2020end}  & $tv$ & $t \to v$ & HT100M & S3D &  & $\checkmark$ & & 15.1 & 38.0 & 51.2 & 10  & \textbf{9.9} & \textbf{24.0} & 32.4 & 29.5 \\
    	\textbf{Ours} & $tva$ & $t \to v$ & HT100M & S3D$\dagger$ & & & $\checkmark$ & 19.8 & 42.9 & 55.1 & 8 & \textbf{9.9} & \textbf{24.0} & \textbf{32.6} & \textbf{28} \\
    	\midrule
    % 	MMT~\cite{gabeur2020multi} & $tva$ & $t \to va$ & HT100M & 7 experts  & $\checkmark$ & & & - & - & - & - & - & 14.4 & - & 66 \\
    	\textbf{Ours} & $tva$ & $t \to va$ & HT100M & S3D$\dagger$ & & & $\checkmark$ & \textbf{24.6} & \textbf{48.3} & \textbf{60.4} & \textbf{6} & 9.3 & 22.9 & 31.2 & 35 \\
    	\midrule
    	\midrule
    % 	MMV \cite{alayrac2020self} & $tva$ & $t \to v$ & HT100M+AudioSet & TSM-50x2 & & $\checkmark$ & $\checkmark$ &11.7 & 33.4 & 45.4 & 13 & 9.3 & 23.0 & 31.1 & 38 \\
    % 	VATT \cite{chen2021multimodal} & $tva$ & $t \to v$ & AudioSet & Transformer & $\checkmark$ & $\checkmark$ & $\checkmark$ & - & - &  45.5 & 13 & - & - & 29.7 & 49 \\
    	FrozenInTime\cite{Bain2021Frozen}& $tv$ &$t \to v$  &  CC+WV+COCO & 
    	Transformer &$\checkmark$ & $\checkmark$ &  & - & - &- & -& 24.7 & 46.9 & 57.2 & 7 \\
    	CLIP4Clip \cite{amrani2020noise} & $tv$ & $t \to v$ & WiT + HT100M & CLIP & $\checkmark$ & $\checkmark$ & & - & - &- & -& 32.0 & 57.0 & 66.9 & 4  \\
    	\bottomrule
    \end{tabular}
    }
    \caption{
    Zero-shot text-to-video retrieval results on YouCook2/MSR-VTT. In ``Retrieval'' column: $v + a$ stands for averaging video and audio embeddings for a video representation, $va$ - our joint video-audio embedding where modalities attend to each other during embedding computation, $t$ and $v$ are single-modality embeddings. S3D$\dagger$ is the S3D pretrained by MIL-NCE~\cite{miech2020end}. We include CLIP4CLIP and FrozenInTime for completeness, but do directly compare because of different pre-training setups. Train Mod.=Training Modalities, BB=Backbone, CC=Conceptual Captions~\cite{sharma2018conceptual}, WV=WedVid-2M~\cite{Bain2021Frozen}.
    \label{tab:retrieval}
    }
    \vspace{-0.25cm}
\end{table*} 
\begin{table*}
    % \tablestyle{3pt}{1.05}

    \resizebox{1\linewidth}{!}{
    \begin{tabular}{@{}l|c|c|c|c|ccc|cccc|cccc@{}}
    	\toprule
    	
        \multirow{2}{*}{Method} & Train. & \multirow{2}{*}{Retrieval} & Pre-train. & Visual & \multicolumn{3}{c}{Trainable BB} & \multicolumn{4}{c}{YouCook2} & \multicolumn{4}{c}{MSR-VTT} \\ 
    	   & Mod. & & Dataset  & BB & ~$t$~ & ~$v$~ & ~$a$~ & R@1$\uparrow$  & R@5$\uparrow$ & R@10$\uparrow$ & MedR$\downarrow$ & R@1$\uparrow$ & R@5$\uparrow$  & R@10$\uparrow$ & MedR$\downarrow$\\ 
    	\midrule
    	ActBERT \cite{zhu2020actbert}  & $tv$ & $t \to v$ & HT100M  & Res3D+Faster R-CNN & & & & - & - & - & - & 16.3 & 42.8 & 56.9 & 10 \\
    	HT100M \cite{miech2019howto100m} & $tv$ & $t \to v$ & ~~~~~~~~HT100M~~~~~~~~ & R152 + RX101 & & & & 8.2 & 24.5 & 35.3 & 24 & 14.9 & 40.2 & 52.8 & 9 \\
    	NoiseEstim. \cite{amrani2020noise}~~~~~~ & $tv$ & $t \to v$ & HT100M & R152 + RX101 & & & & - & - &- & -& 17.4 & 41.6 & 53.6 & 8 \\
    	\upd{\textbf{Ours}} & $tv$ & $t \to v$  & HT100M & R152 + RX101 & & & & 13.7 & 35.3 & 48.4 & 12 & 21.0 & 49.3 & 60.1 & 5 \\
    	\textbf{Ours} & $tva$ & $t \to v$ & HT100M & R152 + RX101 & & & $\checkmark$ & 12.7 & 33.9 & 45.8 & 13 & 20.4 & 47.7 & 59.3 & 6 \\
    	\midrule
    	
    	AVLNet \cite{rouditchenko2020avlnet} & $tva$ & $t \to v+a$ & HT100M & R152 + RX101 & & & $\checkmark$ & 30.2 & 55.5 & 66.5 & 4 & 22.5 & 50.5 & \textbf{64.1} & 5 \\
    	MCN~\cite{chen2021multimodal} & $tva$ & $t \to v+a$ & HT100M & R152 + RX101 & & & $\checkmark$ & 28.2 & 53.0 & 63.7 & 5 & - & - & - & - \\
    	\textbf{Ours} & $tva$ & $t \to va$ & HT100M & R152 + RX101 & & & $\checkmark$ & \textbf{32.1} & \textbf{59.1} & \textbf{70.9} & \textbf{3} & \textbf{23.7} & \textbf{52.1} & 63.7 & \textbf{4}  \\
    	\midrule
    \end{tabular}
    }
    \caption{Text-to-video retrieval on YouCook2/MSR-VTT in the fine-tune setting. In ``Retrieval'' column: $v + a$ stands for averaging video and audio embeddings for a video representation, $va$ - our joint video-audio embedding where modalities attend each other during embedding computation, $t$ and $v$ are single-modality embeddings. Train Mod.=Training Modalities, BB=Backbone.
    \label{tab:sup_retrieval_finetunned}
    }
    % \vspace{-0.3cm}
    % \vspace{-0.25cm}
\end{table*} 

% In this section, we evaluate the performance of the proposed method in two downstream tasks: zero-shot video retrieval and zero-shot step action localization. 
% \noindent Respective code and data will be made publicly available.

%We start with introducing the evaluation setup in Section~\ref{sec:impl_details} and describing the dataset used for self-supervised pretraining and the downstream tasks and datasets used for evaluation in Section~\ref{sec:datasets}. We present an extensive comparison with state-of-the-art methods as well as various ablation studies in Section~\ref{sec:results}. 

% We used narrations without stop words. \info{How stop words were deleted? look at AVLnet paper}.
\subsection{Experimental Setup}
\label{sec:impl_details}

If not stated otherwise, we use the following experimental setup in all our experiments and ablation studies.

\noindent \textbf{Backbones.} To ensure comparability, we follow the setup of previous works~\cite{miech2019howto100m,rouditchenko2020avlnet,chen2021multimodal,amrani2020noise} which is as follows: as visual backbone, we use a combination of ResNet-152~\cite{he2016deep}, pretrained on Imagenet~\cite{deng2009imagenet} and compute one 2D-feature (2048-dimensional vector) per second, as well as ResNeXt-101~\cite{hara2018can} pretrained on Kinetics~\cite{carreira2017quo} to get 1.5 3D-feature (2048 dim.) per second. We temporally upsample 2D-features with nearest neighbors to have the same number of features as 3D-features and concatenate them to obtain 4096-dimensional vectors. As a text backbone, GoogleNews pretrained Word2vec model~\cite{mikolov2013efficient} is used with 300-dimensional embedding per word. These backbones are fixed and not fine-tuned during training. 
Following~\cite{rouditchenko2020avlnet,chen2021multimodal},  we use a trainable CNN with residual layers as an audio backbone %(note that this backbone is not pretrained) 
%that takes log-mel spectrograms  with 16 kHz sampling rate, 25 ms Hamming window, and 10 ms window stride. We 
and adapt the last two residual blocks to extract 1.5 4096-dimensional features per second (see Supplementary Material). 
%For the training, we use a maximum of 20 features for text, 12 features for video and 12 for audio which corresponds to an 8-second video clip. \hkc{perhaps we can condense a bit here ... }
     
\noindent \textbf{Data Sampling.} We use a batch of 224 videos and randomly sample ten 8-second clips per video. 
If the sampled clip contains narration ($95\%$ of all clips), we use ASR time stamps to select clip borders. To disentangle the very high text-audio correlation in HowTo100M, and to avoid text being learned just as an audio narration, we shift the audio clip randomly by 4 seconds with respect to the video and text boundaries.
    
\noindent \textbf{Projections.} Following~\cite{rouditchenko2020avlnet,chen2021multimodal,miech2019howto100m}, we use a gated linear projection~\cite{miech2017learnable} to project features into common token space, as well as to project resulting tokens into shared embedding space. We set the dimension of the common token space to 4096 and of the shared embedding space to 6144. 
    
\noindent \textbf{Transformer architecture.} As a multi-modal fusion transformer, we use one transformer block with a hidden size of 4096, 64 heads, and an MLP size of 4096.
    
\noindent \textbf{Loss computation.} We use a temperature of 0.05 in NCE and normalize vectors prior to computing dot products. Since not every video clip has all three modalities, we computed NCE only over non-empty embeddings. 
Following~\cite{alayrac2020self}, we set a larger weight for a text-visual loss in Equation~\ref{eq:loss}, since it was beneficial for training on HowTo100M: $\lambda_{t\_v} = 1$, $\lambda_{v\_a} = \lambda_{t\_a} = \lambda_{t\_va} = \lambda_{v\_ta} = \lambda_{a\_tv} = 0.1$.
    
\noindent \textbf{Optimization.} We train all models for 15 epochs using an Adam optimizer~\cite{kingma2014adam} with a learning rate of 5e-5 and exponential decay of 0.9.

\subsection{Datasets, Tasks, and Metrics}
\label{sec:datasets}

\noindent \textbf{Pretraining Dataset.} 
We train our model on the HowTo100M dataset~\cite{miech2019howto100m}, which contains over 1 million instructional videos with automatically generated text narrations. The text narrations can be assumed to be noisy and to not always describe the video scene~\cite{miech2019howto100m}. 
% Only $\sim50\%$ videos mention an object or an action that is visually seen in the video clip \cite{miech2019howto100m}.
% We use this dataset to train our model in a self-supervised way as described in Sec.~\ref{sec:loss}. % and during training sampled the 8-second clips as described in Section~\ref{sec:impl_details}.   

%\noindent \textbf{Downstream Tasks, Datasets, and Metrics}
\noindent \textbf{Zero-shot Text-to-video Retrieval.}
We use MSR-VTT~\cite{xu2016msr} and YouCook2~\cite{zhou2018towards} datasets to evaluate the zero-shot text-to-video retrieval capability of our model. 
% which measures how well a model can retrieve video by a text query.
The YouCook2 dataset contains instructional cooking videos from YouTube with human-annotated clips ($\sim2-200$ seconds).
% and a short text descriptions (such as "add oil"). 
For evaluation we use at maximum first 48 seconds of clip, since most clips are shorter than that. The MSR-VTT dataset contains human-annotated video clips ($\sim10-30$ seconds) on various topics 
% (such as "music", "movie", "sport") 
and provides captions with natural language sentences.  Following~\cite{miech2019howto100m,rouditchenko2020avlnet,chen2021multimodal,miech2020end}, to evaluate our model on MSR-VTT, we use the 1k set of test clips~\cite{yu2018joint}, and for YouCook2, we use 3,350 validation clips~\cite{miech2019howto100m}. 
To perform retrieval, we compute similarities by dot product between a text query $t$ and all videos in the dataset using a fused $va$ representation for each video. We report standard recall metrics R@1, R@5, R@10 and the median rank (MedR).
    
\upd{\noindent \textbf{Text-to-video Retrieval after Fine-tuning.}
We additionally evaluate the retrieval performance of the models fine-tuned on downstream tasks. Following~\cite{rouditchenko2020avlnet}, we used 9,586 training clips to fine-tune the model on the YouCook2 dataset, and 6,783 training clips that contain the audio (out of 7,000 proposed by~\cite{miech2019howto100m}) to fine-tune model on the MSR-VTT.}

\noindent \textbf{Zero-shot Step Action Localization.}
We further evaluate our model on zero-shot step action localization tasks on two datasets: CrossTask~\cite{zhukov2019cross} and Mining YouTube~\cite{kuehne2019mining}. The CrossTask dataset consist of 2.7k instructional videos over 18 different tasks.
% such as "Make Lemonade" or "Build Shelves". 
The Mining YouTube dataset provides 250 testing cooking video equipped with an ordered list of action steps.
% and dense per-frame annotation. 
    %Each task has an ordered list of steps with short text descriptions such as "Cut Lemon" and each video corresponds to one task and contains a set of step segments. \hkc{ <- maybe remove that sentence} 
To perform step localization, we use a sliding window and compute the similarity between the current video segment and all step names of the task. Following the inference procedure in~\cite{zhukov2019cross}, we obtain the final labeling by running dynamic programming to find the best labeling with respect to the given order of steps based on similarities. 
\upd{We report average recall over all tasks as defined in~\cite{zhukov2019cross}.}
% We report a recall metric defined as ratio between the number of correct step assignments, where at least one step prediction falls into the ground-truth time interval, and the total number of steps over all videos. We report average recall over all tasks. 
% (each frame is labeled with action class).
% We evaluate our method with the same recall metric as on CrossTask. 
For both datasets, we use a 3-second sliding window with 1-second step and predict the action for the central time-stamp using a fused $va$ representation.

\subsection{Comparison with State-of-the-art}

\label{sec:results}

\noindent \textbf{Zero-shot Text-to-video Retrieval.}
First, we assess the performance of the learned multi-modal representation in context of zero-shot text-to-video retrieval task on YouCook2 and MSR-VTT datasets as shown in Table~\ref{tab:retrieval}.
%For comparison we include only baselines that perform a zero-shot evaluation without tuning the model for downstream tasks for a fair comparison.\hkc{ <- maybe remove that sentence} 
%Since video implicitly comprises two channels, video and audio, for our model, we use a fused video-audio representation $f(v, a)$ for a video. \hkc{ <- maybe remove that sentence} 
In case of the YouCook2 dataset, our method achieves state-of-the-art results over all baselines, including methods with trainable visual and text backbones or a stronger visual backbone as well as methods that do not train the visual backbone. 
Particularly, our approach improves the AVLnet~\cite{rouditchenko2020avlnet} and MCN~\cite{chen2021multimodal} baselines that use the same visual, text, and audio backbone and also train with three modalities by increasing R@10 from 45.2\% to 51.3\%.  For MSR-VTT however, it shows that a fusion of video and audio modalities is not so beneficial and best performance is reached when considering only text to video retrieval and leaving out audio information. We attribute this behaviour to the domain shift between HowTo100M and MSR-VTT datasets as audio of the HowTo100M dataset mainly contains speech and text as a transcription of speech, while in MSR-VTT, audio can be much less related to the textual description. 
This assumption is supported by the fact that best-performing methods on MSR-VTT do not use HowTo100M for training at all, such as FrozenInTime~\cite{Bain2021Frozen} or CLIP4CLIP~\cite{Luo2021CLIP4Clip}.
% which utilizes a strong backbone pre-trained on the WiT dataset CLIP~\cite{Radford2021CLIP} features. 
Notably, we can further strengthen our model on YouCook2 by leveraging a stronger backbone such as S3D~\cite{xie2018rethinking}, pre-trained on HowTo100M by MIL-NCE~\cite{miech2020end}, reaching  R@10 over 60\%. Again, 
% it shows here that 
these results show that
this better adaptation to HowTo100M does not necessarily translate to better results on MSR-VTT. \upd{In the supplement we additionally include experiments with a stronger CLIP~\cite{Radford2021CLIP} backbone.}

\upd{\noindent \textbf{Text-to-video Retrieval after Fine-tuning.} We further evaluated the retrieval performance after fine-tuning on downstream tasks in Table~\ref{tab:sup_retrieval_finetunned}. Note that, since several experimental splits were proposed for the MSR-VTT dataset, we report only baselines that used the same training split as us for a fair comparison.  Results demonstrate that the proposed method clearly outperforms prior works on both datasets. Moreover, after finetuning on the MSR-VTT, the model greatly improves performance by utilizing an audio channel.}
% , achieving a median recall of 3 for the YouCook2 dataset and the median recall of 4 for the MSR-VTT dataset.

\noindent \textbf{Zero-shot Step Action Localization.}
\begin{table}
% 		\tablestyle{2pt}{1.05}]
		\resizebox{\linewidth}{!}{
		\begin{tabular}{@{}l|c|c|c|cc@{}}
            \toprule
            & Tr. & Tr. BB & Visual & \multicolumn{2}{c}{Recall$\uparrow$}\\ 
            % \cmidrule(lr){5-6}
            Method & Mod.& $v$ & Backbone & CrossTask & MYT \\ 
            \midrule
            CrossTask~\cite{zhukov2019cross} & $tv$ & & R152 + I3D & 31.6  & -  \\
            HT100M~\cite{miech2019howto100m} & $tv$ & & R152 + RX101 & 33.6 & 15.0 \\
            MIL-NCE~\cite{miech2020end} & $tv$ & $\checkmark$ & I3D & 36.4 & -  \\
            \midrule
			MCN \cite{chen2021multimodal} & $tva$ & & R152 + RX101 & 35.1 & 18.1 \\
            \textbf{Ours} & $tva$ & & R152 + RX101 &  \textbf{39.3} & \textbf{19.4} \\
            \midrule
            \multicolumn{6}{c}{\upd{Models with a stronger visual backbone:}} \\
            \midrule
            MIL-NCE~\cite{miech2020end}& $tv$ & $\checkmark$ & S3D & 40.5 & - \\
            ActBERT~\cite{zhu2020actbert} & $tv$ & & Res3D+Faster R-CNN  & 41.4 & -  \\
			UniVL \cite{luo2020univilm} & $tv$ & $\checkmark$  &  S3D$\dagger$ & \textbf{42.0} & - \\
			\midrule
			\textbf{Ours} & $tva$ & & S3D$\dagger$ & 41.1 & \textbf{19.7} \\
			\bottomrule
		\end{tabular}
		}
		 \caption{Zero-shot action localization performance on CrossTask/MiningYouTube(MYT). S3D$\dagger$ is the S3D pre-trained by MIL-NCE~\cite{miech2020end}. Tr Mod=Training Modalities, Tr. BB $v$= Trainable Backbone for video modality.
        \label{tab:localization}
		}
        % \vspace{-0.1cm}
\end{table} 
We further evaluate our methods on zero-shot step action localization on the CrossTask and the MiningYouTube (MYT) datasets in Table~\ref{tab:localization}. As video representations we again use the fused video and audio modalities. These results show that the proposed approach clearly outperforms the directly comparable MCN approach on both datasets, as well as a fully supervised baseline CrossTask~\cite{zhukov2019cross}, HT100M~\cite{miech2019howto100m} and MIL-NCE~\cite{miech2020end} with a trainable I3D visual backbone~\cite{miech2020end}. 
\upd{Moreover, with a stronger S3D backbone our model also gains improvement
over MIL-NCE and is comparable to UniVL (with a trainable
backbone) and ActBERT (with additional region-based
features from Faster-R-CNN).}
% Looking at the results for S3D, it shows that the improvement is still there with respect to MIL-NCE, but not sufficient to outperform all other systems, but still comparable to UniVL~\cite{luo2020univilm} and ActBERT~\cite{zhu2020actbert}.

\subsection{Ablation Studies}

\label{sec:ablation}

%In this sections we ablate various design choices of our method and explore the contributions of proposed components.

\begin{table}[]
        \resizebox{\columnwidth}{!}{
		 \begin{tabular}[t]{lccccc}
			    \toprule 
			    \multirow{2}{*}{Configuration} & \multirow{2}{*}{Retrieval} & \multicolumn{2}{c}{YouCook2} & \multicolumn{2}{c}{MSR-VTT}  \\
				&  &  R@5$\uparrow$ & R@10$\uparrow$ & R@5$\uparrow$ & R@10$\uparrow$ \\
				\hline
				% AVLnet* & $t \to v+a$ & 32.1 & 41.7 & 23.9 & 32.7 \\
				1) no transformers & $t \to v+a$ & 32.7 & 41.4 & 24.1 & 33.7 \\
				2) single-mod. transformer per mod. & $t \to v+a$& 39.9 & 50.7 & 25.3 & 33.9 \\
				3) fusion transformer & $t \to v+a$ & 39.5 & 50.2 & 23.8 & 32.7 \\
				4) fusion transformer & $t \to va$ & 36.6 & 47.0 & 22.6 & 32.1 \\
				5) fusion transf. + comb. loss & $t \to v+a$ & 38.2 & 49.2 & 23.3 & 33.2 \\
				6) fusion transf. + comb. loss (ours) & $t \to va$ & 40.7 & 51.3 & 23.8 & 31.8 \\
			\bottomrule
		\end{tabular}
		}
		\caption{Evaluation of the contribution of the proposed fusion transformer and the combinatorial loss.  In ``Retrieval'' column: $v + a$ stands for independently extracting video and audio embeddings and summing up both outputs, while $va$  for forwarding both modalities together allowing them to attend to each other. 
        \label{tab:fusion}
		}
		\vspace{-0.3cm}
\end{table}

\noindent \textbf{Impact of fusion components.} 
%In Table~\ref{tab:fusion} we evaluate our multi-modal fusion transformer and the combinatorial loss to answer a question \textit{which setup is better to learn a joint video and audio embedding: fusion or summing.} 
We first address the question of how the proposed components: transformer layer, transformer fusion, and combinatorial loss, impact the overall performance of our system. 
To this end, in Table~\ref{tab:fusion} we considered the following architectures: 
%1) \textit{AVLnet*} as a baseline trained in our experimental setup (data sampling, batch size, loss, etc. as described in Section~\ref{sec:impl_details}); 
1) \textit{no transformers}: our architecture without transformer 
and with three pairwise contrastive losses;
2) \textit{single modality transformer}: using three separate modality-specific transformer layers to learn three projection functions; 
3) \textit{fusion transformer}: using the proposed modality agnostic transformer 
% with combinatorial input,
but trained with three pairwise contrastive losses without fused modality components; %in this setup, the model architecture is the same as ours and allow us to fuse tokens from different modalities, but it was not trained to do it. 
4) \textit{fusion transformer + comb. loss}: using the proposed modality agnostic transformer with combinatorial input, trained with combinatorial loss to obtain the proposed method. Schematic visualization of these four setups is included in the supplement. We further consider two ways to forward two modalities, first by forwarding them separately and summing up both outputs $(v + a)$ and, second, by forwarding them together $(va)$.
Overall we observe that adding a simple transformer to process each modality separately already significantly improves performance compared to the baseline, especially for the YouCook2 dataset.
%
%Considering the performance of the proposed modality agnostic transformer as well as the combinatorial loss, 
We further observe that the overall performance depends on the combination of model, loss function and fusion strategy at test time.
While token fusion with transformers is overall beneficial, the best performance is achieved in the \textit{fusion transformer + comb. loss} setup. When using \textit{fusion transformer} alone, the performance drops a bit  compared to the \textit{single modality transformer}. However, utilizing shared transformer with combinatorial loss to fuse tokens from different modalities outperforms modality \textit{summing} in the \textit{single modality transformer} setup.

\begin{table}[]
        \resizebox{\columnwidth}{!}{
		 \begin{tabular}[t]{lcccc}
			    \toprule 
			    \multicolumn{1}{c}{} & \multicolumn{2}{c}{YouCook2} & \multicolumn{2}{c}{MSR-VTT}  \\
				Configuration  &  R@5$\uparrow$ & R@10$\uparrow$ & R@5$\uparrow$ & R@10$\uparrow$ \\
				\hline
				% ours & 41.7 & 52.6 & 24.2 & 33 \\
				ours & 40.7 & 51.3 & 23.8 & 31.8 \\
				ours + shared final proj. & 37.2 & 48.7 & 23.1 & 29.4 \\
				ours + $[cls]$ token + shared final proj. & 35.9 & 46.9 & 21.4 & 28.8  \\
			%\midrule
			\bottomrule
		\end{tabular}
		}
		\vspace{-0.2cm}
		\caption{Evaluation of different design choices for fusion transformer (with/without $[cls]$ token) and final projection (shared/modality specific) into multi-modal embedding space.
        \label{tab:token_agg_and_proj}
		}
		\vspace{-0.3cm}
\end{table}

\noindent \textbf{Token Aggregation and Projection.}
We further evaluate the impact of separate processing of output tokens compared to aggregating information in the  $[cls]$ token. To this end, we compare our model architecture to a setup with a shared final projection, as well as a setup with an additional $[cls]$ input token (similarly to BERT~\cite{devlin2018bert}).
% that serves as a joint representation of all tokens. 
In the last scenario, we use the output $[cls]$ of the multi-modal fusion transformer as an aggregated representation of input tokens and apply the final shared projection to map it into the shared embedding space. At shown in Table~\ref{tab:token_agg_and_proj}, our modality-specific projections with no $[cls]$ token benefit over others options on both datasets.

\begin{table}[]
        \resizebox{\columnwidth}{!}{
		 \begin{tabular}[t]{lcccc}
			    \toprule 
			    \multicolumn{1}{c}{} & \multicolumn{2}{c}{YouCook2} & \multicolumn{2}{c}{MSR-VTT}  \\
				Configuration  &  R@5$\uparrow$ & R@10$\uparrow$ & R@5$\uparrow$ & R@10$\uparrow$ \\
				\hline
				positional emb. + uniform sampling & 32.7 & 43.4 & 21.8 & 29.1\\
				positional emb. + max polling & 33.2 & 43.9 & 21.6 & 30.5   \\
				positional emb. + averaging over clips & 35.6 & 46.9 & 23.0 & 29.2 \\
				\midrule
				no positional emb. + uniform sampling & 34.5 & 45.1 &  22.1 & 30.2   \\
				no positional emb. + max polling & 34.8 & 45.2 &   23.0 & 31.8 \\
				no positional emb. + averaging over clips & 36.9 & 47.9 & 22.7 & 31.3 \\
				no positional emb. + video at once &  40.7 & 51.3 & 23.8 & 31.8\\
			\bottomrule
		\end{tabular}
		}
		\vspace{-0.2cm}
		\caption{Evaluation of different strategies to process long videos (longer than clips used in the training), as well as impact of positional embeddings to encode positional information.  
        \label{tab:testing_longer_clips}
        }
        \vspace{-0.3cm}
\end{table}

\noindent \textbf{Positional Embedding and Testing on Longer Clips.}
Finally, we address the question how the option of having random length inputs impacts the overall performance of the model at test time. We consider four different scenarios for testing on longer clips as shown in Table~\ref{tab:testing_longer_clips}: 1) \textit{uniform sampling} - after obtaining initial local features from backbones, features are uniformly sampled to fit the maximum number of tokens; 2) \textit{max-pooling} - local features are merged via adaptive max-pooling; 3) \textit{averaging over clips} comprises slicing a longer clip into train-time-length clips and averaging obtained representations;  4) \textit{video at once} considers processing all features at once as proposed. We further compare the first three settings in a scenario with positional embedding to no positional embedding. As positional embedding, we used vanilla trainable embeddings~\cite{devlin2018bert} that are summed up to the input tokens before being input to the transformer. 
It shows that setups with positional embedding perform almost consistently below the setups without positional embedding on both datasets.
% and that this observation holds for both datasets, YouCook2 and mostly MSR-VTT.
% Looking at results of different processing strategies, we find that our model benefits from leveraging all context via \textit{max-pooling}, \textit{uniform sampling}, or \textit{video at once} during embedding computation compared to slicing a video clip into shorter clips. Moreover, utilizing all input data at \textit{video at once} further boosts performance. 
%
Looking at results of different processing strategies, we find that our model benefits from leveraging local temporal dependencies in data via slicing a video clip into shorter clips or \textit{video at once} compared to \textit{max-pooling}, \textit{uniform sampling}. Moreover, utilizing all input data at \textit{video at once} further significantly boosts performance. 

\begin{figure}[]
    \centering 
    \begin{subfigure}{0.65\linewidth}
        \includegraphics[width=\textwidth]{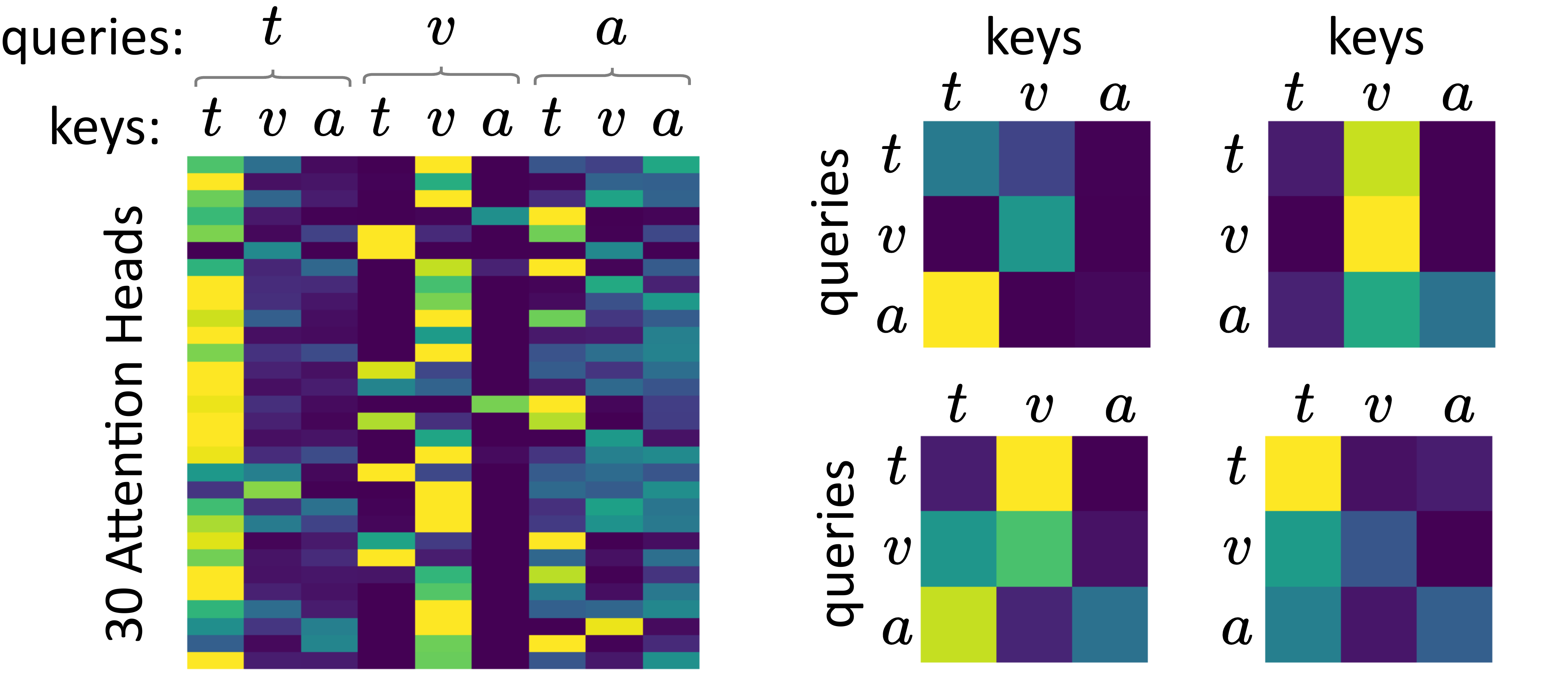}
        % \captionof{figure}{}
    \end{subfigure}
    % \vfill
    \vspace{-0.25cm}
    \caption{Attention Analysis. Left: Average attentions of 30 heads over 128 video clips for queries and keys from different  modalities. Right: average attention for random 4 heads presented in 3x3 matrix.
    \label{fig:attention}}
    \vspace{-0.5cm}
\end{figure}

\noindent \textbf{Attention Analysis.}
Finally, we qualitatively analyze the fusion capability of our multi-modal transformer. In Figure~\ref{fig:attention} we show the average attention for query-key pairs of tokens from different modalities. We observe that some heads have a strong attention for single-modality fusion, mostly for $t$ and $v$ modalities, and in between, some heads are responsible for cross-modal attention.

\vspace{-0.1cm}
\section{Limitations and Conclusion}
In this work, we propose a multi-modal, modality agnostic fusion transformer approach that learns to exchange information between multiple modalities, such as video, audio, and text, and to integrate them into a joined multi-modal representation.
% to obtain an embedding that aggregates multi-modal temporal information. 
\upd{We show that training the system with a combinatorial loss on any possible combinations of modalities allows the fusion transformer to learn a strong multi-modal embedding space leaving out any add-ons such as position encoding.} 
% We show that training the system with a combinatorial loss on single modalities as well as pairs of modalities allows the fusion transformer to learn possible combinations of input modalities 
% that leaving out any add-ons such as position encoding can improve performance as well. 
A clear limitation of the system becomes evident when looking at the performance difference on two downstream datasets, YouCook2 and MSR-VTT, showing that a better fusion can result in a loss in generalizability to multi-modal data that was acquired in a different way. 
A future research direction to mitigate those effects might be to consider techniques from domain adaptation or generalization in context of multi-modal zero-shot recognition. We hope that the proposed setup might inspire further research on this topic as well as on self-attention based multi-modal video processing in general. 
\vspace{-0.1cm}
\section{Acknowledgments}

This work was supported by the Goethe Center for Scientific Computing (G-CSC) at Goethe University Frankfurt, the MIT-IBM-Satori team, and the IBM-MIT Watson AI Lab.

%%%%%%%%% REFERENCES
{\small
\bibliographystyle{ieee_fullname}
\bibliography{egbib}
}

\clearpage
\appendix
\addcontentsline{toc}{section}{Appendix} 
\part{Supplementary Material} % Start the appendix part
{\hypersetup{linkcolor=black} \parttoc}

% \numberwithin{equation}{section}
% \setcounter{table}{0}
% \setcounter{figure}{0}
% \setcounter{equation}{0}

\noindent Supplementary material is organized as follows: first, we provide additional experimental results in Section~\ref{sec:add_exps}; then, we perform a qualitative analysis of zero-shot text-to-video retrieval in Section~\ref{sec:qualitative}; and finally, we provide more implementation details in Section~\ref{sec:more_imp}.

\section{Additional Experimental Evaluation}
\label{sec:add_exps}
% \subsection{Stronger Visual Backbone}
\upd{\subsection{CLIP Backbone} 
We additionally tested our model with stronger visual and text backbones. Namely, we used CLIP backbones (Contrastive Language-Image Pre-training)~\cite{radford2021learning} pre-trained on the large Wikipedia-based image-text WiT dataset. We used the ViT-B/32 model and extracted one 512-dimensional feature per second for video and one 512-dimensional feature per word for text. For both modalities, we adopt features after projection into the multi-modal embedding space.
Performance of zero-shot text-to-video retrieval and text-to-video retrieval after fine-tuning is presented in Table~\ref{tab:CLIP}. We note that using CLIP features is especially beneficial for the MSR-VTT dataset, but performance on YouCook2 also improves compared to R152 + RX101 and word2vec backbones. We also note that performance on MSR-VTT after fine-tuning is coming close to the performance of the CLIP4CLIP~\cite{Luo2021CLIP4Clip} model that, however, is not directly comparable to ours. Compared to CLIP4CLIP, we are not fine-tuning backbones and we also are using a smaller MSR-VTT train subset for fine-tuning (7,000 clips compared to 9,000 clips).}

% We evaluated CLIP features in Table~\ref{tab:CLIP}. We found CLIP features especially beneficial on MSR-VTT after fine-tuning (see Section A.3 of suppl.).}

% \input{tables/supplement/retrieval}

% We also tested our model with a stronger visual backbone. Namely, we used S3D features pre-trained on HowTo100M by MIL-NCE~\cite{miech2020end} or CLIP features (Contrastive Language-Image Pre-training)~\cite{radford2021learning} pre-trained on the large Wikipedia-based image-text WiT dataset, as well as a combination of them. For S3D, we extract 1024-dimensional features at 16 fps in contiguous chunks of 16 frames, resulting in 1 feature per second. For CLIP, we used the ViT-B/32 model and extracted 1 512-dimensional feature per second (we adopt features after projection into multi-modal embedding space). Note that in all our experiments, we keep GoogleNews pre-trained Word2vec model~\cite{mikolov2013efficient}
% as a text backbone and trainable CNN with residual connection as an audio backbone. 

% Performance of zero-shot text-to-video retrieval on YouCook2 and MSR-VTT datasets with different visual backbones is presented in Table~\ref{tab:sup_retrieval}. We note that using CLIP visual features is especially beneficial for the MSR-VTT dataset, while performance on YouCook improves only slightly compared to S3D features. 
% \todo{Do we really need it?}

\subsection{Action Segmentation}

Following~\cite{chen2021multimodal} we additionally report temporal action segmentation performance on the CrossTask and Mining YouTube datasets as proposed in~\cite{kuehne2019mining}. We measured a frame-wise video segmentation performance given the order of actions in a video. Following inference procedure~\cite{kuehne2019mining} we computed temporal alignment of video frames based on similarity matrix to text labels by a Viterbi-decoding. Before decoding, we transferred the similarity matrix to class probabilities by applying softmax with temperature 0.05 across all labels over all videos (as we did in NCE during training). Segmentation performance is measured by an intersection over union $IoU = \frac{G \cap D}{G \cup D}$  -- the ratio between the intersection of ground truth action $G$ and prediction $D$ and the union of them -- as well as an intersection over detection $IoD = \frac{G \cap D}{D}$. 

In Table~\ref{tab:localization_iou} we show $IoU$ and $IoD$ for temporal action segmentation with a recall for step action localization. We observe that our method shows a marginal boost in temporal action segmentation on the Mining YouTube dataset while it does not benefit on the CrossTask dataset. However, we note that the segmentation evaluation procedure relies on the given order of steps in a video, while in the CrossTask dataset about 30\% steps are missed and step orders are not always correct~\cite{zhukov2019cross}, so we consider the step action localization recall as a primary metric for this dataset, where our method improves performance by 4\% with respect to MCN~\cite{chen2021multimodal} baseline. 

\begin{table}[t]
        % \vspace{-0.2cm}
        \resizebox{\columnwidth}{!}{
		 \begin{tabular}[t]{lllccccc}
			    \toprule 
			    Method & Visual & Text & \multirow{2}{*}{FT} & \multicolumn{2}{c}{YouCook2} & \multicolumn{2}{c}{MSR-VTT}  \\
				 & Backbone & Backbone & & R@5$\uparrow$ & R@10$\uparrow$ & R@5$\uparrow$ & R@10$\uparrow$ \\
				\hline
				Ours & R152+RX101 & word2vec &  &  40.7 & 51.3 & 23.8 & 31.8  \\
				Ours & CLIP & word2vec &  & 42.7 & 54.0 & 29.0 & 38.7  \\
			    Ours & CLIP & CLIP & & 42.6 & 54.3 & 32.5 & 42.4 \\
			    \midrule
			    CLIP4CLIP & CLIP & CLIP & & - & - & 57.0 & 66.9 \\
			    \midrule
			    \midrule
			    Ours &R152+RX101 & word2vec & $\checkmark$ & 59.1 & 70.9 & 52.1 & 63.7  \\
			    Ours &CLIP & word2vec & $\checkmark$ & 62.1 & 72.6 & 60.7 & 72.7  \\
			    Ours & CLIP & CLIP & $\checkmark$ & 62.1 & 72.9 & 62.7 & 75.0 \\
			    \midrule
			    CLIP4CLIP & CLIP & CLIP & $\checkmark$ & - & - & 70.7 & 80.5 \\
			\bottomrule 
		\end{tabular}
		}
% 		\vspace{-0.35cm}
		\caption{Text-to-video retrieval on the YouCook2/MSR-VTT in zero-shot and fine-tune settings with CLIP backbones. As the video representation, we again use $va$ -- the fused video and audio modalities. FT: fine-tuning on downstream task. We include CLIP4CLIP~\cite{Luo2021CLIP4Clip} for completeness but do directly compare because of different pre-training and a different MSR-VTT train subset.
		\label{tab:CLIP}}
        % \vspace{-0.35cm}
\end{table}
\begin{table}
% 		\tablestyle{2pt}{1.05}]
		\resizebox{\linewidth}{!}{
		\begin{tabular}{@{}l|ccc|ccc@{}}
            \toprule
            & \multicolumn{3}{c}{CrossTask} & \multicolumn{3}{c}{Mining YouTube} \\
            Method & Recall$\uparrow$ & IOD$\uparrow$ & IOU$\uparrow$ & Recall$\uparrow$ & IOD$\uparrow$ & IOU$\uparrow$  \\ 
            \midrule
            Mining YouTube~\cite{kuehne2019mining} & - & - & - & - & 19.2 &  9.8 \\
 			MCN \cite{chen2021multimodal} & 35.1 & \textbf{33.6} & \textbf{22.2} & 18.1 & 32.0 & \textbf{23.1} \\
            \textbf{Ours} & \textbf{39.3} & 32.5 & 18.5 & \textbf{19.4} & \textbf{32.7} & \textbf{23.1} \\
			\bottomrule
		\end{tabular}
		}
		 \caption{Evaluation of zero-shot action segmentation on the CrossTask/Mining YouTube. We report results for ``R152 + RX101'' visual backbone (the same as used in MCN~\cite{chen2021multimodal}). 
        \label{tab:localization_iou}
		}
        % \vspace{-0.1cm}
\end{table} 
\begin{table*}[h]
\centering
\resizebox{1\textwidth}{!}{
\begin{tabular}{@{}l|c@{~~~~}c@{~~}c@{~~}c@{~~}c@{~~}c@{~~}c@{~~}c@{~~}c@{~~}c@{~~}c@{~~}c@{~~}c@{~~}c@{~~}c@{~~}c@{~~}c@{~~}c@{~~}|c@{}} \toprule
    Method & \rotatebox{90}{\small Make} \rotatebox{90}{\small Kimchi Rice}  
    & \rotatebox{90}{\small Pickle} \rotatebox{90}{\small Cucumber}  
    & \rotatebox{90}{\small Make Banana} \rotatebox{90}{\small Ice Cream}  
    & \rotatebox{90}{\small Grill} \rotatebox{90}{\small Steak}  
    & \rotatebox{90}{\small Jack Up } \rotatebox{90}{\small Car}  
    & \rotatebox{90}{\small Make } \rotatebox{90}{\small Jello Shots}  
    & \rotatebox{90}{\small Change } \rotatebox{90}{\small Tire}  
    & \rotatebox{90}{\small Make } \rotatebox{90}{\small Lemonade}  
    & \rotatebox{90}{\small Add Oil } \rotatebox{90}{\small to Car}  
    & \rotatebox{90}{\small Make } \rotatebox{90}{\small Latte}  
    & \rotatebox{90}{\small Build } \rotatebox{90}{\small Shelves}  
    & \rotatebox{90}{\small Make } \rotatebox{90}{\small Taco Salad}  
    & \rotatebox{90}{\small Make } \rotatebox{90}{\small French Toast}  
    & \rotatebox{90}{\small Make } \rotatebox{90}{\small Irish Coffee}  
    & \rotatebox{90}{\small Make } \rotatebox{90}{\small Strawberry Cake}  
    & \rotatebox{90}{\small Make } \rotatebox{90}{\small Pancakes}  
    & \rotatebox{90}{\small Make } \rotatebox{90}{\small Meringue}  
    & \rotatebox{90}{\small Make } \rotatebox{90}{\small Fish Curry}  
    & \rotatebox{90}{\small Average }
\\ \midrule

Fully-supervised baseline~\cite{zhukov2019cross}                           & 19.1          & 25.3          & 38.0          & 37.5          & 25.7          & 28.2          & \textbf{54.3}          & 25.8          & 18.3         & 31.2          & \textbf{47.7}          & 12.0          & 39.5          & 23.4          & 30.9          & 41.1          & 53.4          & 17.3          & 31.6 \\ 
\midrule
CrossTask \cite{zhukov2019cross}                         & 13.3          & 18.0          & 23.4 & 23.1 & 16.9          & 16.5 & 30.7 & 21.6 & 4.6          & 19.5 & 35.3        & 10.0 & 32.3 & 13.8 & 29.5 & 37.6 & {43.0} & 13.3 & 22.4 \\
HT100M \etal ~\cite{miech2019howto100m}                    & \textbf{33.5}           & {27.1}         & {36.6}           & 37.9           & {24.1}           & 35.6 & {32.7}           & {35.1}           & \textbf{30.7}         & {28.5}           & {43.2}         & {19.8}           & {34.7}           & {33.6}           & 40.4           & {41.6}          & 41.9           & {27.4}           & {33.6} \\
MCN & 25.5 &    31.1 &    39.7 &    32.7 &    35.4 &    \textbf{36.8} &    29.0 &    40.0 &    28.4 &    \textbf{33.8} &    45.7 &    27.5 &    36.1 &    34.9 &    39.6 &    42.6 &    43.0 &    29.1 & 35.1 \\
\textbf{Ours}& 30.5 &  \textbf{41.2} &   \textbf{46.5} &   \textbf{46.6}  &  \textbf{38.9} & 32.0 &    19.5 &   \textbf{48.9} &   25.8 &   33.6 &   44.7  &  \textbf{29.1} &  \textbf{40.7}  &   \textbf{36.9} &    \textbf{50.7}  &    \textbf{44.1} &   \textbf{63.1} &   \textbf{33.6} & \textbf{39.2} \\

\bottomrule
\end{tabular}
}
\vspace{-0.1cm}
\caption{Step action localization performance on the CrossTask~\cite{zhukov2019cross} dataset: recalls corresponding to every specific task.}
\vspace{-0.1cm}
\label{tab:action_step_localization}
\end{table*}
\begin{table}[]
    \resizebox{\columnwidth}{!}{
		 \begin{tabular}[t]{cccc|cc|cc}
			    \toprule 
			    \multirow{2}{*}{\#blocks} & \multirow{2}{*}{\#heads} & \multirow{2}{*}{hidden s.} & \multirow{2}{*}{batch size}& \multicolumn{2}{c}{YouCook2} & \multicolumn{2}{c}{MSR-VTT}  \\
				 & & &   &  R@5$\uparrow$ & R@10$\uparrow$ & R@5$\uparrow$ & R@10$\uparrow$ \\
				\midrule
				1 & 64 & 4096 & $224 \times 10$ & \textbf{40.7} & \textbf{51.3} & \textbf{23.8} & 31.8 \\
				2 & 64 & 4096 & $224 \times 5$ & 37.3 & 47.6 & 23.2 & \textbf{32.5} \\
				2 & 64 & 4096 & $112 \times 10$ & 38.6 & 49.8 & 20.8 & 28.6 \\
				2 & 32 & 2048 & $224 \times 10$ & 38.1 & 49.1 & 22.6 & 30.9 \\
				4 & 16 & 1024 & $224 \times 10$ & 35.4 & 46.8 & 23.7 & 31.7 \\
			\bottomrule
		\end{tabular}
		}
		\caption{Evaluation of different fusion transformer architectures. \textit{\#blocks} stands for a number of transformer blocks,  \textit{\#heads} -- for a count of attention heads; \textit{hidden s.} denotes a hidden size of the transformer layers (that linearly depends on the number of heads); \textit{batch size} denotes a training batch size where $x \times y$ means that we use a batch of $x$ videos and randomly sample $y$ clips per video.
        \label{tab:abl_trans_conf}
		}
		
\end{table}

\begin{table}[]
    \resizebox{\columnwidth}{!}{
		 \begin{tabular}[t]{l|cc|cc}
			    \toprule 
			    \multirow{2}{*}{Configuration} & \multicolumn{2}{c}{YouCook2} & \multicolumn{2}{c}{MSR-VTT}  \\
				  &  R@5$\uparrow$ & R@10$\uparrow$ & R@5$\uparrow$ & R@10$\uparrow$ \\
				\midrule
				aligned text-audio & 37.8 & 47.2 & 14.0 & 20.0 \\
				disentangled text-audio & 37.2 & 45.7 & 17.9 & 25.0 \\ 
				disentangled text-audio + loss weighting & \textbf{40.7} & \textbf{51.3} & \textbf{23.8} & \textbf{31.8} \\
			\bottomrule
		\end{tabular}
		}
  \vspace{-0.1cm}
		\caption{Evaluation of disentangling of audio and text while training on the  HowTo100M dataset as well weighting components in the loss function. \textit{Aligned text-audio} and \textit{disentangled text-audio} were trained without loss weighting, \textit{disentangled text-audio + loss weighting} --  with $\lambda_{t\_v} = 1$, $\lambda_{v\_a} = \lambda_{t\_a} = \lambda_{t\_va} = \lambda_{v\_ta} = \lambda_{a\_tv} = 0.1$ as proposed. 
        \label{tab:audio_sampling}
		}
  \vspace{-0.1cm}
		
\end{table}

\subsection{CrossTask Specific Results}

To further analyze step action localization performance, we considered recalls for every specific task of the CrossTask dataset in Table~\ref{tab:action_step_localization}. We note that our method shows a significant boost in almost all cooking-related categories, like ``Make Banana Ice Cream'' or ``Grill Steak'' while does not improve performance in not-cooking categories ``Change Tire,'' ``All Oil to Car,'' and ``Build a Shelves.'' The MCN method, which also utilizes audio channel, similarly demonstrate a lower performance in ``Change Tire'' and ``All Oil to Car'' tasks compared to video-text-only the CrossTask~\cite{zhukov2019cross} and HT100M~\cite{miech2019howto100m} baselines. We can assume that this happens due to the fewer car-related video clips in the HowTo100M dataset (7.8M) compared to food-related clips (54.4 M).

% \subsection{Finetunning}
% \input{tables/supplement/finetunned}

% For text-to-video retrieval, we additionally include the performance of the models fine-tuned on downstream tasks. Following~\cite{rouditchenko2020avlnet}, we used 9,586 training clips to tune model on the YouCook2 dataset, and $\sim$7,000 video clips (proposed by~\cite{miech2019howto100m}) to fine-tune model on the MSR-VTT dataset. Following~\cite{rouditchenko2020avlnet}, we used only 6,783 clips that contain the audio during fine-tuning on the MSR-VTT. Note that, since several experimental splits were proposed for the MSR-VTT dataset, we report only baselines that used the same training split as us for a fair comparison.  Results presented in Table~\ref{tab:sup_retrieval_finetunned} demonstrate that the proposed method outperforms prior works, achieving a median recall of 3 for the YouCook2 dataset and the median recall of 4 for the MSR-VTT dataset. 

\subsection{Fusion Transformer Ablation}

We also additionally ablate our Fusion Transformer with respect to the number of transformer blocks and the number of heads of multi-headed attention (and the hidden size of the transformer layer) in Table~\ref{tab:abl_trans_conf}. Due to resource constraints, for an increase in the number of transformer blocks, we should linearly decrease either the number of heads or the training batch size (the large batch size is essential due to contrastive training). We observe that the best performing configuration consists of 1 transformer block and a maximum number of transformer heads (64 heads) that fits into resources, however, we assume the model can further boost performance by increasing the number of transformer blocks leveraging more resources.

\subsection{Text-Audio Disentangling and Loss Weighting}

We also show the importance of disentangling audio with respect to text while training on the HowTo100M dataset, as well as the importance of using a larger text-video loss weight in the loss function in Table~\ref{tab:audio_sampling}. Since text is obtained by applying an ASR system to the audio track, to avoid text being learned just as an audio narration, we shift the audio clip randomly by half of clip length (4 seconds out of 8 seconds) with respect to the video and text boundaries in \textit{``disentangled text-audio.''} To further regularize text-audio learning, in \textit{``+ loss weighting''} we used a larger weight for a text-visual loss $\lambda_{t\_v} = 1$ compared to other loss components $\lambda_{v\_a} = \lambda_{t\_a} = \lambda_{t\_va} = \lambda_{v\_ta} = \lambda_{a\_tv} = 0.1$ (similarly to~\cite{alayrac2020self}). Table~\ref{tab:audio_sampling} shows that both adaptations are beneficial for training on the HowTo100M dataset. 

 \begin{figure*}
    \centering
    
    \begin{subfigure}[t]{0.95\textwidth}
        \begin{subfigure}[t]{0.16\textwidth}
        \centering
        \stackinset{c}{0pt}{c}{0pt}{\Centerstack{
        \footnotesize \textbf{Text Query}  
        }}{\includegraphics[width=0.16\linewidth]{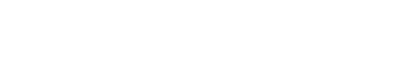}}
        \end{subfigure}%
        \begin{subfigure}[t]{0.82\textwidth}
        \centering
        \stackinset{c}{0pt}{c}{0pt}{\Centerstack{
        \footnotesize \textbf{Top 5 Retrieved Videos} 
        }}{\includegraphics[width=0.16\linewidth]{images/retrieval/empty_short.png}}
        \end{subfigure}
        
        \begin{subfigure}[t]{0.16\textwidth}
        \centering
        \stackinset{c}{0pt}{c}{0pt}{\Centerstack{
        \footnotesize heat the oil and \\
        \footnotesize fry the falafel balls \\
        \footnotesize until golden brown
        }}{\includegraphics[width=\textwidth]{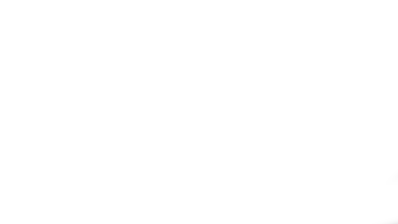}}
         \end{subfigure}
         \begin{subfigure}[t]{0.16\textwidth}
             \centering
            \includegraphics[width=\textwidth]{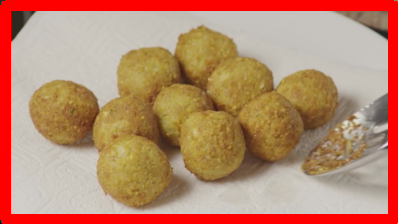}
         \end{subfigure}
         \begin{subfigure}[t]{0.16\textwidth}
             \centering
            \includegraphics[width=\textwidth]{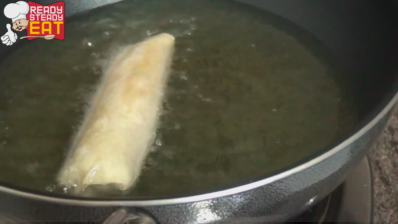}
         \end{subfigure}
         \begin{subfigure}[t]{0.16\textwidth}
             \centering
            \includegraphics[width=\textwidth]{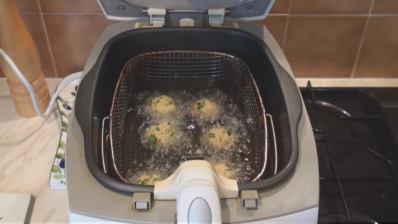}
         \end{subfigure}
         \begin{subfigure}[t]{0.16\textwidth}
             \centering
            \includegraphics[width=\textwidth]{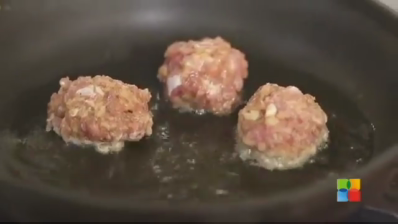}
         \end{subfigure}
         \begin{subfigure}[t]{0.16\textwidth}
             \centering
            \includegraphics[width=\textwidth]{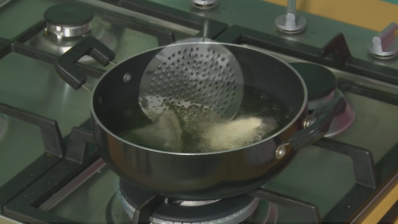}
         \end{subfigure}

         \begin{subfigure}[t]{0.16\textwidth}
        \centering
        \stackinset{c}{0pt}{c}{0pt}{\Centerstack{
        \footnotesize fold the foil \\
        \footnotesize around the sandwich
        }}{\includegraphics[width=\textwidth]{images/retrieval/empty.png}}
         \end{subfigure}
         \begin{subfigure}[t]{0.16\textwidth}
             \centering
            \includegraphics[width=\textwidth]{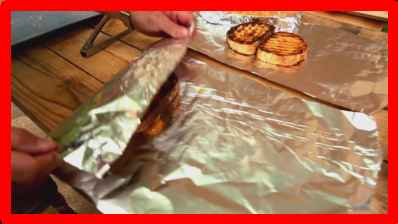}
         \end{subfigure}
         \begin{subfigure}[t]{0.16\textwidth}
             \centering
            \includegraphics[width=\textwidth]{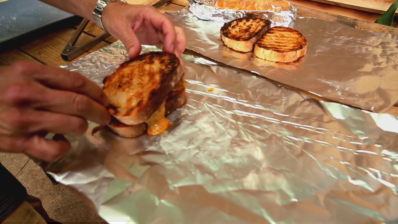}
         \end{subfigure}
         \begin{subfigure}[t]{0.16\textwidth}
             \centering
            \includegraphics[width=\textwidth]{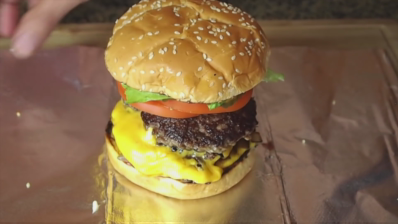}
         \end{subfigure}
         \begin{subfigure}[t]{0.16\textwidth}
             \centering
            \includegraphics[width=\textwidth]{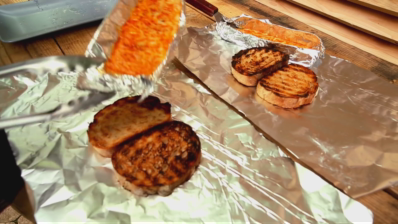}
         \end{subfigure}
         \begin{subfigure}[t]{0.16\textwidth}
             \centering
            \includegraphics[width=\textwidth]{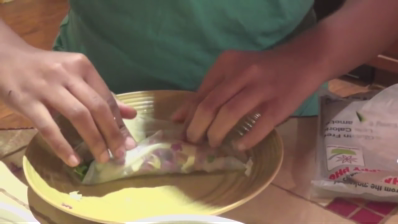}
         \end{subfigure}

         \begin{subfigure}[t]{0.16\textwidth}
        \centering
        \stackinset{c}{0pt}{c}{0pt}{\Centerstack{
        \footnotesize cover meat with   \\
        \footnotesize flour dunk in eggs \\ 
        \footnotesize and coat in bread crumbs
        }}{\includegraphics[width=\textwidth]{images/retrieval/empty.png}}
         \end{subfigure}
         \begin{subfigure}[t]{0.16\textwidth}
             \centering
            \includegraphics[width=\textwidth]{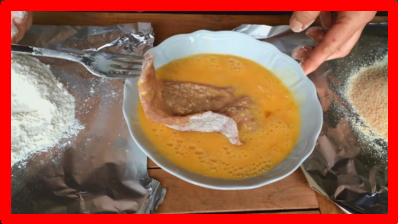}
         \end{subfigure}
         \begin{subfigure}[t]{0.16\textwidth}
             \centering
            \includegraphics[width=\textwidth]{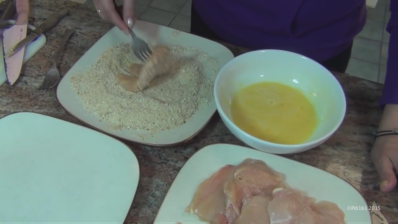}
         \end{subfigure}
         \begin{subfigure}[t]{0.16\textwidth}
             \centering
            \includegraphics[width=\textwidth]{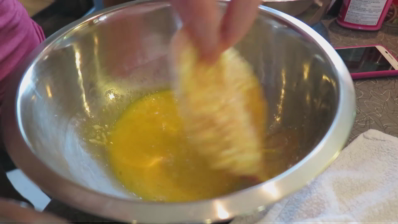}
         \end{subfigure}
         \begin{subfigure}[t]{0.16\textwidth}
             \centering
            \includegraphics[width=\textwidth]{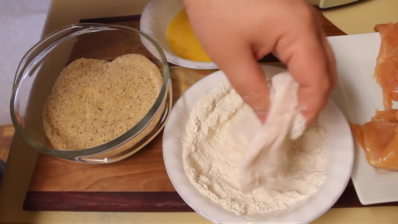}
         \end{subfigure}
         \begin{subfigure}[t]{0.16\textwidth}
             \centering
            \includegraphics[width=\textwidth]{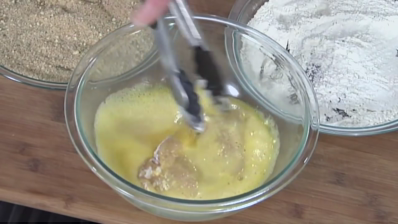}
         \end{subfigure}

         \begin{subfigure}[t]{0.16\textwidth}
        \centering
        \stackinset{c}{0pt}{c}{0pt}{\Centerstack{
        \footnotesize mix the yeast sugar   \\
        \footnotesize and water
        }}{\includegraphics[width=\textwidth]{images/retrieval/empty.png}}
         \end{subfigure}
         \begin{subfigure}[t]{0.16\textwidth}
             \centering
            \includegraphics[width=\textwidth]{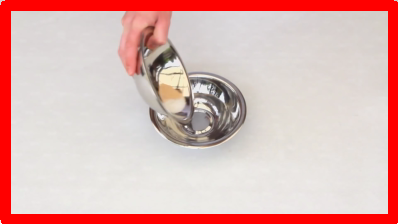}
         \end{subfigure}
         \begin{subfigure}[t]{0.16\textwidth}
             \centering
            \includegraphics[width=\textwidth]{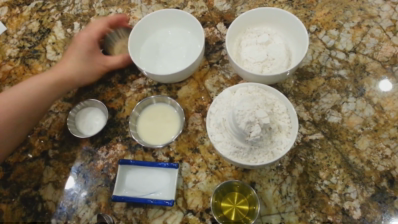}
         \end{subfigure}
         \begin{subfigure}[t]{0.16\textwidth}
             \centering
            \includegraphics[width=\textwidth]{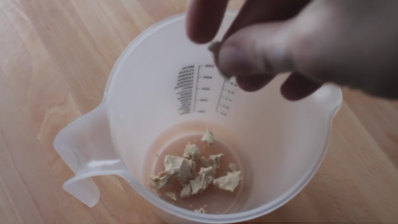}
         \end{subfigure}
         \begin{subfigure}[t]{0.16\textwidth}
             \centering
            \includegraphics[width=\textwidth]{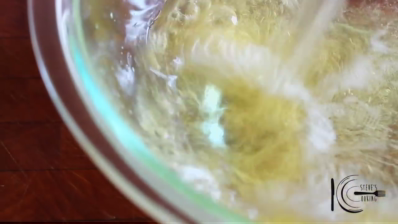}
         \end{subfigure}
         \begin{subfigure}[t]{0.16\textwidth}
             \centering
            \includegraphics[width=\textwidth]{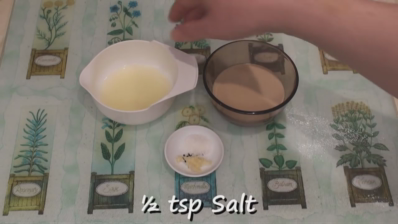}
         \end{subfigure}
         
         \caption{Examples of clips retrieved in the top-1 results ($@R = 1$)}
    \end{subfigure}

    \begin{subfigure}[t]{0.95\textwidth}
        \begin{subfigure}[t]{0.16\textwidth}
        \centering
        \stackinset{c}{0pt}{c}{0pt}{\Centerstack{
        \footnotesize \textbf{Text Query}  
        }}{\includegraphics[width=0.16\linewidth]{images/retrieval/empty_short.png}}
        \end{subfigure}%
        \begin{subfigure}[t]{0.82\textwidth}
        \centering
        \stackinset{c}{0pt}{c}{0pt}{\Centerstack{
        \footnotesize \textbf{Top 5 Retrieved Videos} 
        }}{\includegraphics[width=0.16\linewidth]{images/retrieval/empty_short.png}}
        \end{subfigure}
        
        \begin{subfigure}[t]{0.16\textwidth}
        \centering
        \stackinset{c}{0pt}{c}{0pt}{\Centerstack{
        \footnotesize cook the macaroni  \\
        \footnotesize in boiling water
        }}{\includegraphics[width=\textwidth]{images/retrieval/empty.png}}
         \end{subfigure}
         \begin{subfigure}[t]{0.16\textwidth}
             \centering
            \includegraphics[width=\textwidth]{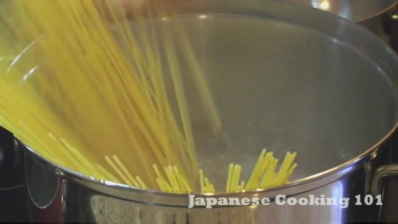}
         \end{subfigure}
         \begin{subfigure}[t]{0.16\textwidth}
             \centering
            \includegraphics[width=\textwidth]{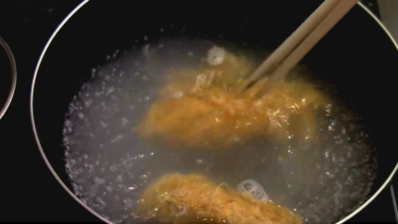}
         \end{subfigure}
         \begin{subfigure}[t]{0.16\textwidth}
             \centering
            \includegraphics[width=\textwidth]{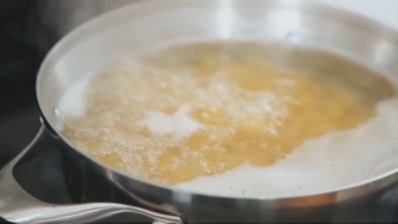}
         \end{subfigure}
         \begin{subfigure}[t]{0.16\textwidth}
             \centering
            \includegraphics[width=\textwidth]{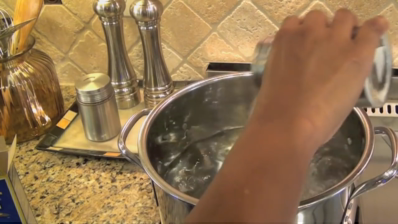}
         \end{subfigure}
         \begin{subfigure}[t]{0.16\textwidth}
             \centering
            \includegraphics[width=\textwidth]{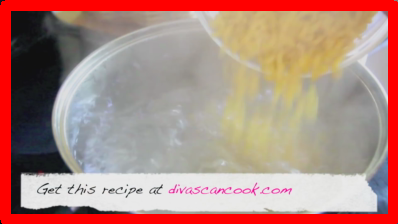}
         \end{subfigure}
         
         \begin{subfigure}[t]{0.16\textwidth}
        \centering
        \stackinset{c}{0pt}{c}{0pt}{\Centerstack{
        \footnotesize spray the chicken \\
        \footnotesize with cooking spray \\
        \footnotesize and cook the chicken \\
        \footnotesize in the oven
        }}{\includegraphics[width=\textwidth]{images/retrieval/empty.png}}
         \end{subfigure}
         \begin{subfigure}[t]{0.16\textwidth}
             \centering
            \includegraphics[width=\textwidth]{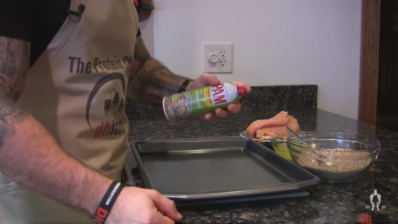}
         \end{subfigure}
         \begin{subfigure}[t]{0.16\textwidth}
             \centering
            \includegraphics[width=\textwidth]{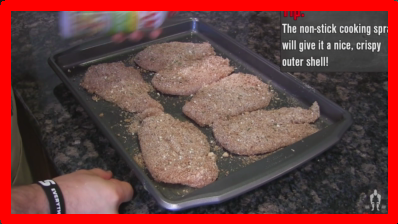}
         \end{subfigure}
         \begin{subfigure}[t]{0.16\textwidth}
             \centering
            \includegraphics[width=\textwidth]{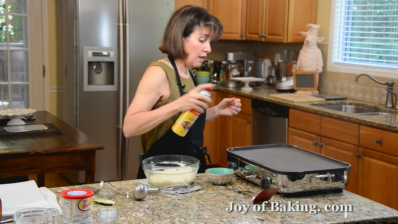}
         \end{subfigure}
         \begin{subfigure}[t]{0.16\textwidth}
             \centering
            \includegraphics[width=\textwidth]{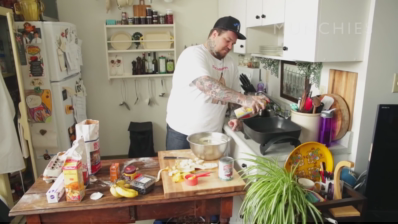}
         \end{subfigure}
         \begin{subfigure}[t]{0.16\textwidth}
             \centering
            \includegraphics[width=\textwidth]{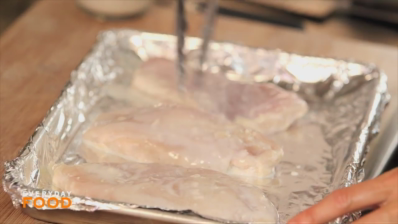}
         \end{subfigure}
         
        \begin{subfigure}[t]{0.16\textwidth}
        \centering
        \stackinset{c}{0pt}{c}{0pt}{\Centerstack{
        \footnotesize combine diced tomato \\
        \footnotesize and cucumber and \\
        \footnotesize sliced onions
        }}{\includegraphics[width=\textwidth]{images/retrieval/empty.png}}
         \end{subfigure}
         \begin{subfigure}[t]{0.16\textwidth}
             \centering
            \includegraphics[width=\textwidth]{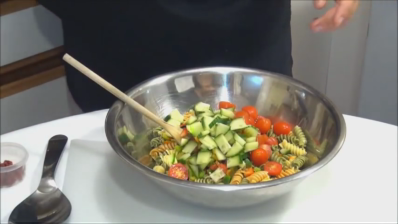}
         \end{subfigure}
         \begin{subfigure}[t]{0.16\textwidth}
             \centering
            \includegraphics[width=\textwidth]{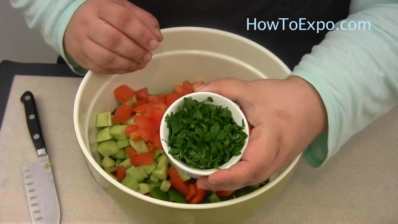}
         \end{subfigure}
         \begin{subfigure}[t]{0.16\textwidth}
             \centering
            \includegraphics[width=\textwidth]{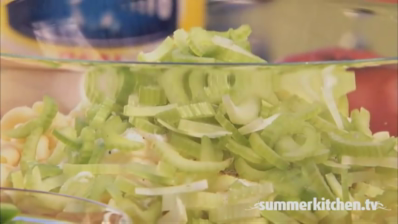}
         \end{subfigure}
         \begin{subfigure}[t]{0.16\textwidth}
             \centering
            \includegraphics[width=\textwidth]{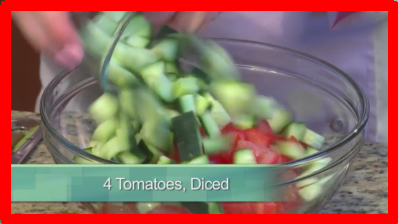}
         \end{subfigure}
         \begin{subfigure}[t]{0.16\textwidth}
             \centering
            \includegraphics[width=\textwidth]{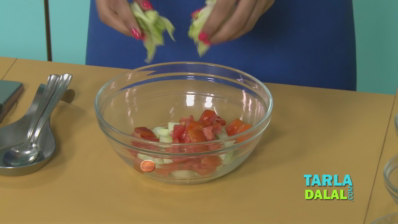}
         \end{subfigure}

         \begin{subfigure}[t]{0.16\textwidth}
        \centering
        \stackinset{c}{0pt}{c}{0pt}{\Centerstack{
        \footnotesize chop some fresh \\
        \footnotesize parsley 
        }}{\includegraphics[width=\textwidth]{images/retrieval/empty.png}}
         \end{subfigure}
         \begin{subfigure}[t]{0.16\textwidth}
             \centering
            \includegraphics[width=\textwidth]{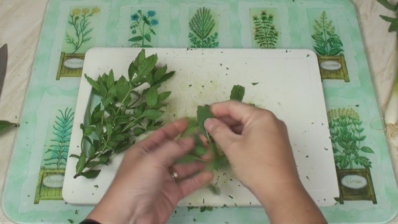}
         \end{subfigure}
          \begin{subfigure}[t]{0.16\textwidth}
             \centering
            \includegraphics[width=\textwidth]{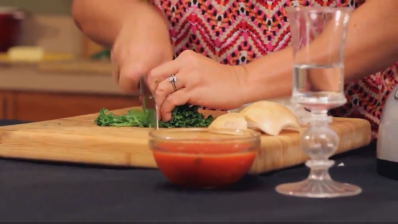}
         \end{subfigure}
          \begin{subfigure}[t]{0.16\textwidth}
             \centering
            \includegraphics[width=\textwidth]{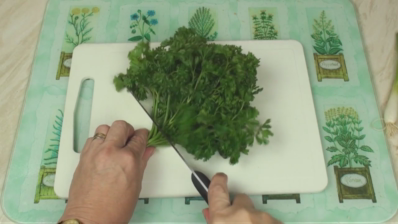}
         \end{subfigure}
          \begin{subfigure}[t]{0.16\textwidth}
             \centering
            \includegraphics[width=\textwidth]{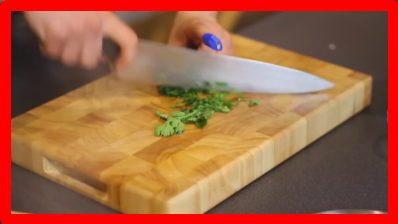}
         \end{subfigure}
          \begin{subfigure}[t]{0.16\textwidth}
             \centering
            \includegraphics[width=\textwidth]{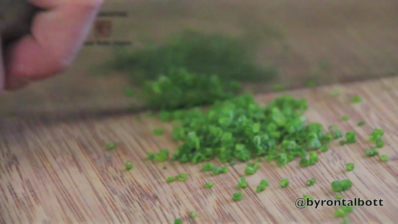}
         \end{subfigure}
         
         \caption{Examples of clips retrieved in the top-5 results ($@R <= 5$)}
    \end{subfigure}

    \begin{subfigure}[t]{0.95\textwidth}
        \begin{subfigure}[t]{0.16\textwidth}
        \centering
        \stackinset{c}{0pt}{c}{0pt}{\Centerstack{
        \footnotesize \textbf{Text Query}  
        }}{\includegraphics[width=0.16\linewidth]{images/retrieval/empty_short.png}}
        \end{subfigure}%
        \begin{subfigure}[t]{0.82\textwidth}
        \centering
        \stackinset{c}{0pt}{c}{0pt}{\Centerstack{
        \footnotesize \textbf{Top 5 Retrieved Videos}
        }}{\includegraphics[width=0.16\linewidth]{images/retrieval/empty_short.png}}
        \end{subfigure}
        
        \begin{subfigure}[t]{0.16\textwidth}
        \centering
        \stackinset{c}{0pt}{c}{0pt}{\Centerstack{
        \footnotesize bring a large pan  \\
        \footnotesize of water to boil
        }}{\includegraphics[width=\textwidth]{images/retrieval/empty.png}}
         \end{subfigure}
         \begin{subfigure}[t]{0.16\textwidth}
             \centering
            \includegraphics[width=\textwidth]{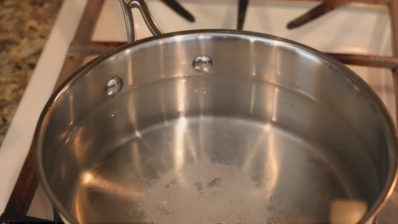}
         \end{subfigure}
         \begin{subfigure}[t]{0.16\textwidth}
             \centering
             \includegraphics[width=\textwidth]{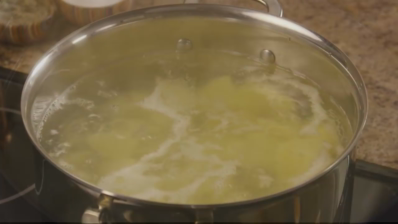}
         \end{subfigure}
         \begin{subfigure}[t]{0.16\textwidth}
             \centering
             \includegraphics[width=\textwidth]{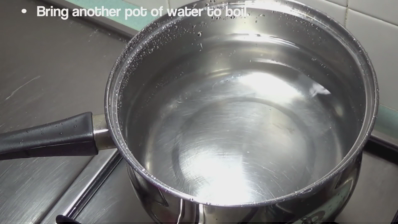}
         \end{subfigure}
         \begin{subfigure}[t]{0.16\textwidth}
             \centering
             \includegraphics[width=\textwidth]{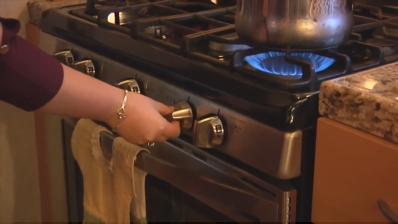}
         \end{subfigure}
         \begin{subfigure}[t]{0.16\textwidth}
             \centering
             \includegraphics[width=\textwidth]{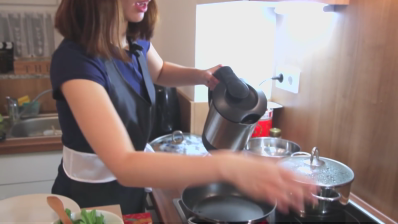}
         \end{subfigure}
         
        \begin{subfigure}[t]{0.16\textwidth}
        \centering
        \stackinset{c}{0pt}{c}{0pt}{\Centerstack{
        \footnotesize when air bubbles form  \\
        \footnotesize flip the bread over
        }}{\includegraphics[width=\textwidth]{images/retrieval/empty.png}}
         \end{subfigure}
         \begin{subfigure}[t]{0.16\textwidth}
             \centering
            \includegraphics[width=\textwidth]{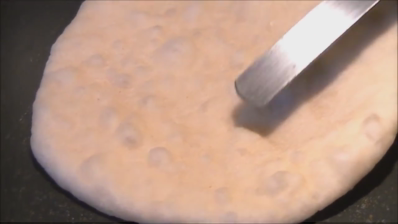}
         \end{subfigure}
         \begin{subfigure}[t]{0.16\textwidth}
             \centering
            \includegraphics[width=\textwidth]{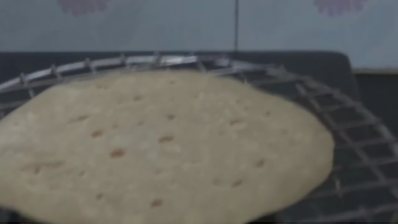}
         \end{subfigure}
         \begin{subfigure}[t]{0.16\textwidth}
             \centering
            \includegraphics[width=\textwidth]{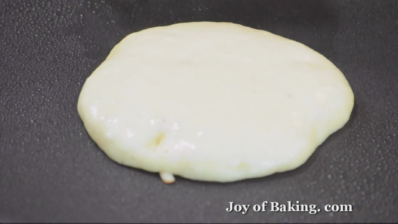}
         \end{subfigure}
         \begin{subfigure}[t]{0.16\textwidth}
             \centering
            \includegraphics[width=\textwidth]{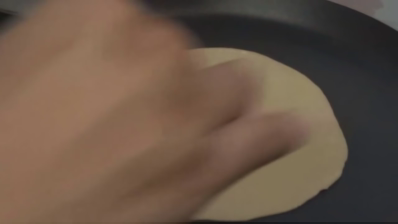}
         \end{subfigure}
         \begin{subfigure}[t]{0.16\textwidth}
             \centering
            \includegraphics[width=\textwidth]{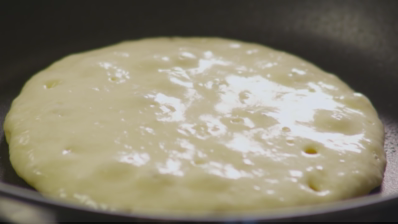}
         \end{subfigure}
         
        \begin{subfigure}[t]{0.16\textwidth}
        \centering
        \stackinset{c}{0pt}{c}{0pt}{\Centerstack{
        \footnotesize add worcestershire sauce \\
        \footnotesize to the pot
        }}{\includegraphics[width=\textwidth]{images/retrieval/empty.png}}
         \end{subfigure}
         \begin{subfigure}[t]{0.16\textwidth}
             \centering
            \includegraphics[width=\textwidth]{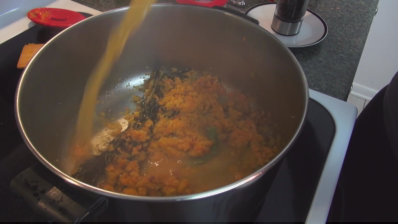}
         \end{subfigure}
         \begin{subfigure}[t]{0.16\textwidth}
             \centering
            \includegraphics[width=\textwidth]{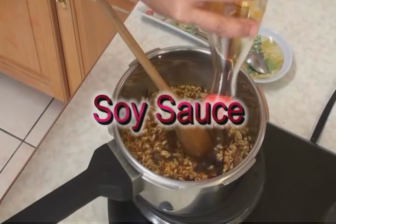}
         \end{subfigure}
         \begin{subfigure}[t]{0.16\textwidth}
             \centering
            \includegraphics[width=\textwidth]{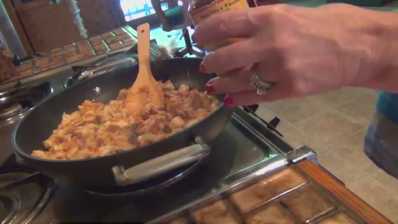}
         \end{subfigure}
         \begin{subfigure}[t]{0.16\textwidth}
             \centering
            \includegraphics[width=\textwidth]{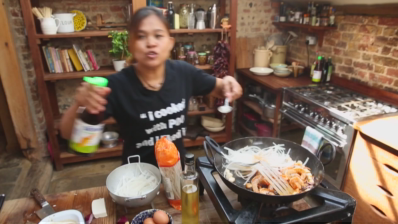}
         \end{subfigure}
         \begin{subfigure}[t]{0.16\textwidth}
             \centering
            \includegraphics[width=\textwidth]{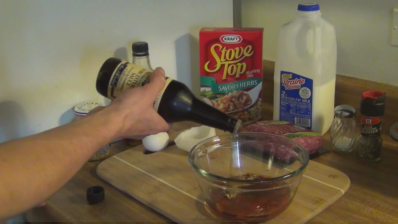}
         \end{subfigure}
         
         \begin{subfigure}[t]{0.16\textwidth}
        \centering
        \stackinset{c}{0pt}{c}{0pt}{\Centerstack{
        \footnotesize take out the wrapped \\
        \footnotesize ingredients add boiling \\
        \footnotesize water and cover \\
        \footnotesize the jar
        }}{\includegraphics[width=\textwidth]{images/retrieval/empty.png}}
         \end{subfigure}
         \begin{subfigure}[t]{0.16\textwidth}
             \centering
            \includegraphics[width=\textwidth]{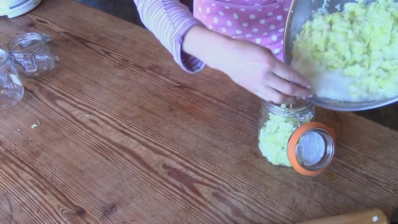}
         \end{subfigure}
         \begin{subfigure}[t]{0.16\textwidth}
             \centering
            \includegraphics[width=\textwidth]{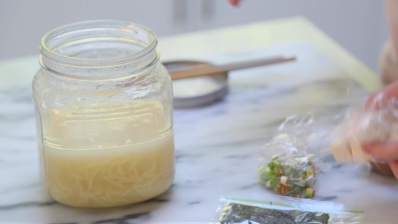}
         \end{subfigure}
         \begin{subfigure}[t]{0.16\textwidth}
             \centering
            \includegraphics[width=\textwidth]{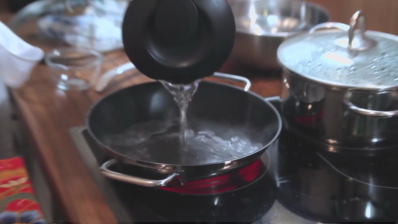}
         \end{subfigure}
         \begin{subfigure}[t]{0.16\textwidth}
             \centering
            \includegraphics[width=\textwidth]{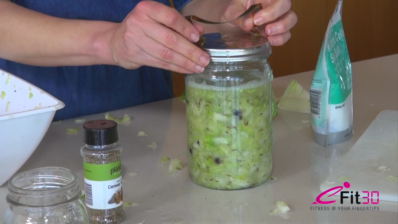}
         \end{subfigure}
         \begin{subfigure}[t]{0.16\textwidth}
             \centering
            \includegraphics[width=\textwidth]{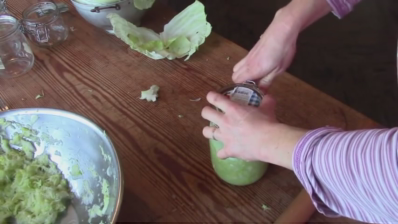}
         \end{subfigure}
         \caption{Examples of clips not retrieved in the top-5 results ($@R > 5$).}

    \end{subfigure}

     \caption{Qualitative evaluation. Examples of zero-shot text-to-video retrieval on the YouCook2. Each row shows the top-5 retrieved videos for a given text query. The correct video is highlighted with a red color.}
    \label{fig:retrieval}

\end{figure*}

\begin{figure*}
    \centering 
    % \vspace{0.3cm}
    \begin{subfigure}{0.8\linewidth}
        \includegraphics[width=\textwidth]{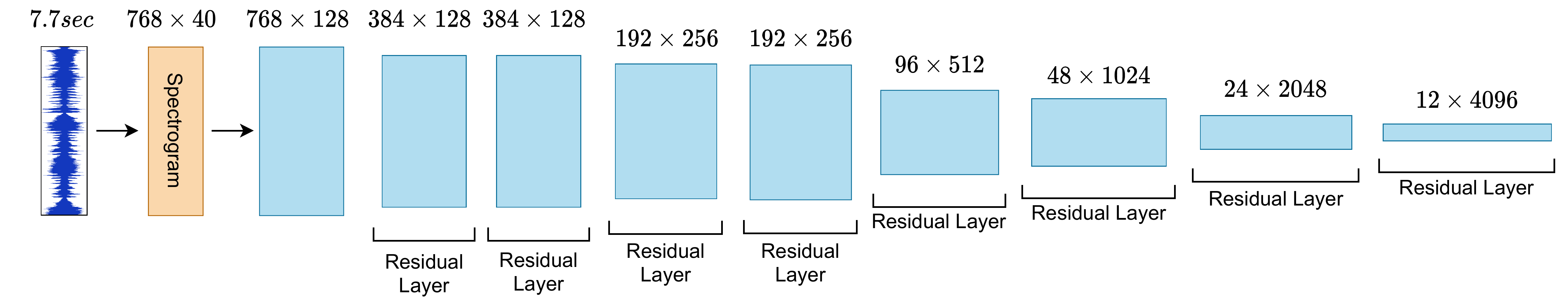}
        % \captionof{figure}{}
    \end{subfigure}
    % \vfill
    \caption{The schematic visualization of audio backbone network (the illustration is inspired by~\cite{rouditchenko2020avlnet}).
    \label{fig:audio_backbone}}
    % \vspace{-0.5cm}
\end{figure*}

\begin{figure*}
    \centering 
    % \vspace{0.3cm}
    \begin{subfigure}{0.48\linewidth}
        \includegraphics[width=\textwidth]{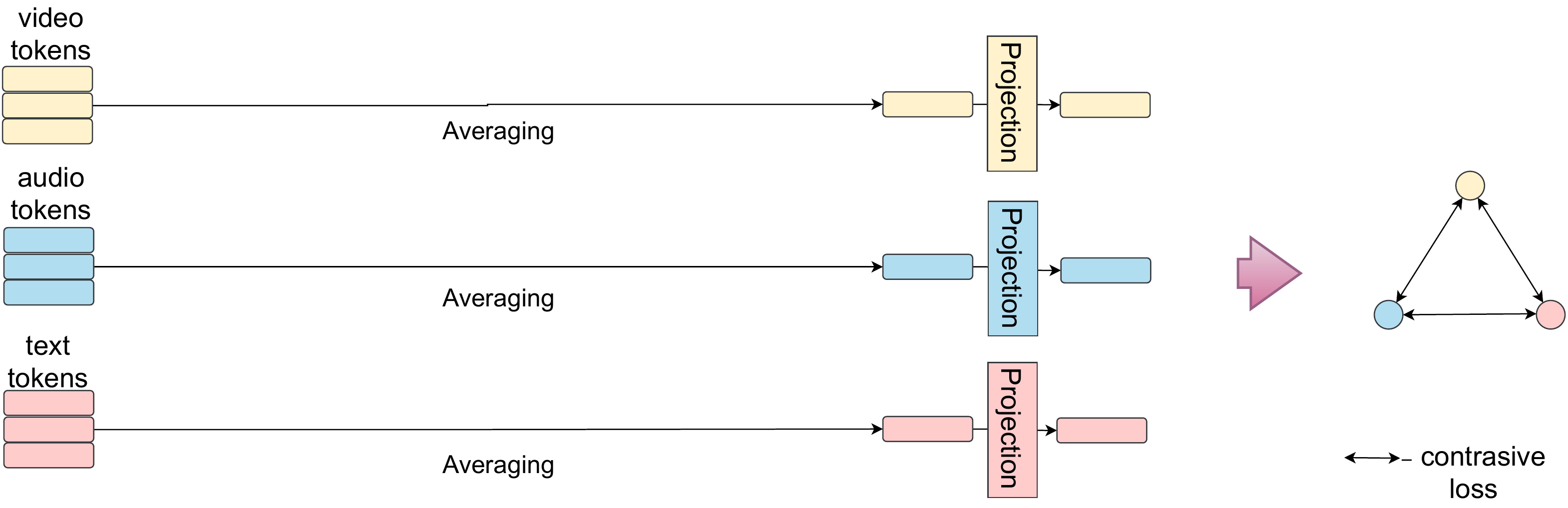}
        \caption{\textit{No Transformers}}
        % \captionof{figure}{}
    \end{subfigure}%
    \hfill
    \begin{subfigure}{0.48\linewidth}
        \includegraphics[width=\textwidth]{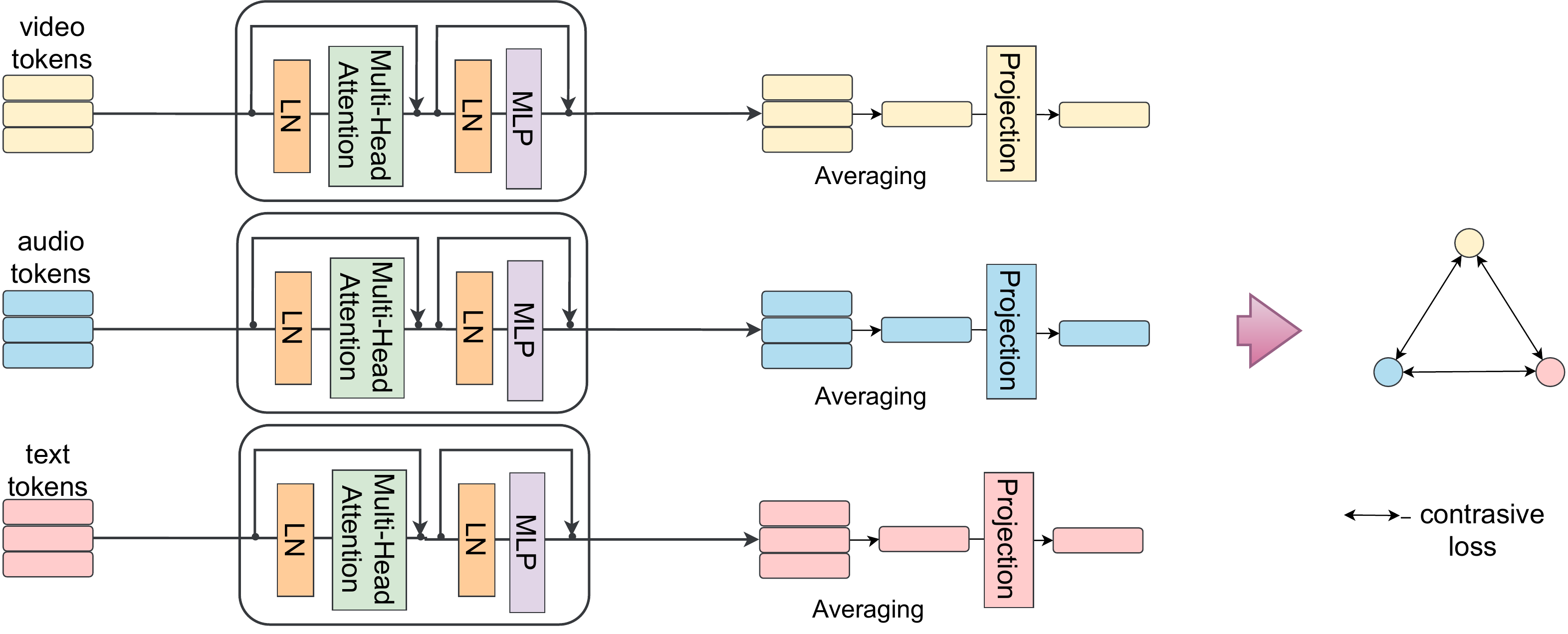}
        \caption{\textit{Single Modality Transformer}}
        % \captionof{figure}{}
    \end{subfigure}
    
    \begin{subfigure}{0.48\linewidth}
        \includegraphics[width=\textwidth]{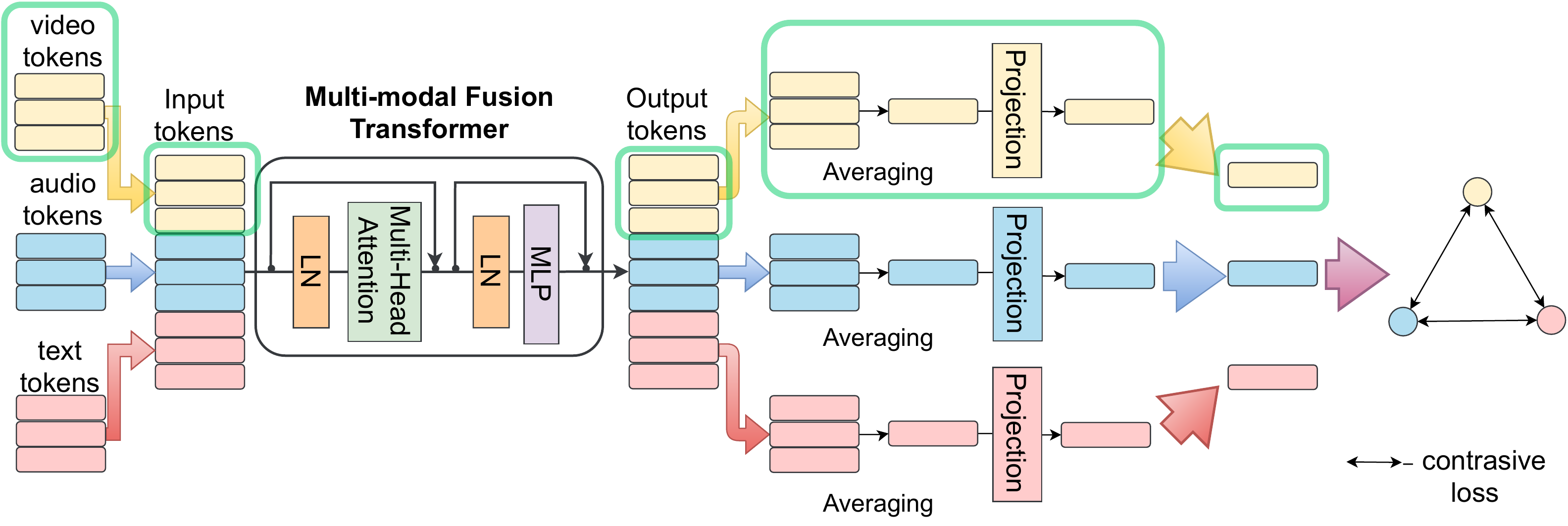}
        \caption{\textit{Fusion Transformer}}
        % \captionof{figure}{}
    \end{subfigure}%
    \hfill
    \begin{subfigure}{0.48\linewidth}
        \includegraphics[width=\textwidth]{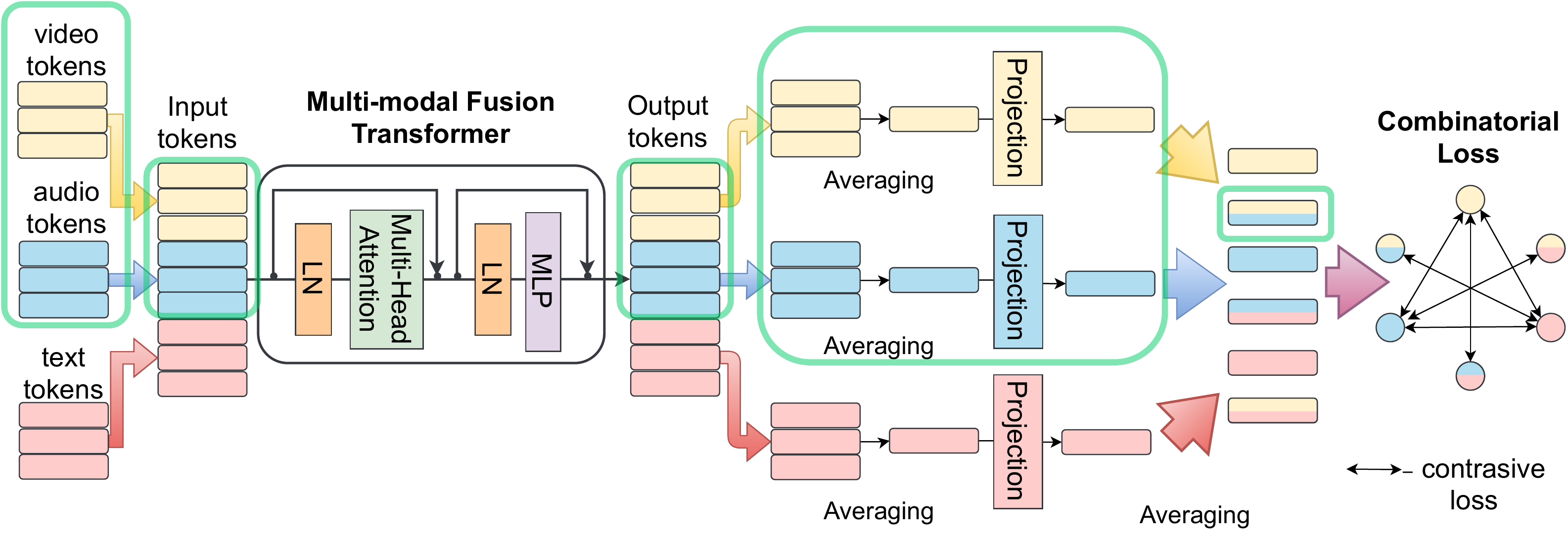}
        \caption{\textit{Fusion Transformer + Combinatorial Loss}
        \label{fig:abl_archs_ours}
        }
        % \captionof{figure}{}
    \end{subfigure}
    
    % \vfill
    \caption{Comparison of different architectures considered in ablation studies. Note that in illustration of the fusion transformer in (c) and (d), not all blocks are always active, using green rectangles we consider the video embedding computation in (c) and video-audio embedding computation in (d). 
    \label{fig:abl_archs}}
    % \vspace{-0.5cm}
\end{figure*}

\subsection{Relative Positional Encodings} 

\begin{table}[t]
% 		\vspace{-0.3cm}
        \centering
        \resizebox{0.95\columnwidth}{!}{
		 \begin{tabular}[t]{lcccc}
			    \toprule 
			    \multicolumn{1}{c}{} & \multicolumn{2}{c}{YouCook2} & \multicolumn{2}{c}{MSR-VTT}  \\
				Configuration  &  R@5$\uparrow$ & R@10$\uparrow$ & R@5$\uparrow$ & R@10$\uparrow$ \\
				\midrule
			    RoPE~\cite{su2021roformer} & 40.4 & 51.2 & 23.1 & 31.6 \\
				no positional emb. &  \textbf{40.7} & \textbf{51.3} & \textbf{23.8} & \textbf{31.8}  \\
			\bottomrule
		\end{tabular}
		}
		\vspace{-0.1cm}
		\caption{Evaluation of Rotary Position Embedding (RoPE).
		\label{tab:RoPE}}
            \vspace{-0.1cm}
        % \vspace{-0.3cm}
\end{table}

\upd{As demonstrated in the paper, we found that absolute positional embeddings are not beneficial for our model. Apart from them, we also tested relative positional encodings, namely the Rotary Position Embedding (RoPE)~\cite{su2021roformer}, that are
shown to better generalize to longer inputs at test time. We incorporated RoPE into our attention block, where we independently apply RoPE to each sequence of text, video, or audio tokens. However, as presented in Table~\ref{tab:RoPE} we also found that RoPE does not benefit our model. But a more complex strategy that e.g., adds information about alignment tokens from different modalities (similarly to the RoPE 2D case) may lead to performance improvement.}

\section{Qualitative Analysis}

\label{sec:qualitative}

We also qualitatively analyze the zero-shot retrieval capacity of our model on the YouCook2 dataset in~Figure~\ref{fig:retrieval}. We observe, that for all shown examples retrieved clips are semantically related to the given text query. Even when a correct video does not occur in the top-5 retrieval results, top-5 videos correspond to the text input: for example, for a query ``bring a large pan of water to boil'' our model predicts videos with boiling water in a pot. 

\section{Implementation Details}

\label{sec:more_imp}

\subsection{Audio Backbone}

Following~\cite{rouditchenko2020avlnet,chen2021multimodal}, as an audio backbone, we use a trainable CNN with residual layers adopted from~\cite{harwath2018jointly} that takes log-mel spectrograms with 16 kHz sampling rate, 25 ms Hamming window,  10 ms window stride, and 40 Mel filter bands. Note that this backbone is not pretrained. Since architecture used in ~\cite{rouditchenko2020avlnet,chen2021multimodal} extracts 6 1024-dimensional features per second, we adapt the last two residual blocks to extract $\sim$1.5 4096-dimensional features per second (the same as our video backbone). We illustrate architecture in Figure~\ref{fig:audio_backbone}. While training on 8-seconds clips, we used 7.7 seconds of audio, that results exactly in 12 audio tokens.

\subsection{Ablation Architectures}

In Figure~\ref{fig:abl_archs} we illustrate 4 architectures considered in our ablation studies: 
a) \textit{no transformers}: our architecture without transformer layer, trained with three pairwise contrastive losses;
2) \textit{single modality transformer}: leveraging three separate modality-specific transformers; 
3) \textit{fusion transformer}: the proposed modality agnostic transformer, but trained with three pairwise contrastive losses without fused modality components; 
4) \textit{fusion transformer + combinatorial loss}: the proposed architecture that utilises the modality agnostic transformer with combinatorial input, trained with combinatorial loss.

\subsection{Fine-tuning Details}

During fine-tuning on the YouCook2 and MST-VTT datasets, we set $\lambda_{t\_v} = \lambda_{v\_a} = \lambda_{t\_a} = \lambda_{t\_va} = \lambda_{v\_ta} = \lambda_{a\_tv} = 1$, and train the model for 5 epochs with a learning rate of $1e^{-5}$ and a batch size of 256 on the YouCook2 dataset, and for 25 epochs with the learning rate of $5e^{-5}$ and the batch size of 128 on the MSR-VTT dataset. 

\subsection{Training Time}

Training our model on the HowTo100M dataset takes approximately 2 days on four Nvidia V100 32GB GPUs. Fine-tuning on the YouCook2 and the MSR-VTT takes less than 30 minutes. 
% --- supplementary material
% \input{sec/X_supplementary}

\end{document}